\documentclass[runningheads]{llncs}

\usepackage{makecell}
\usepackage{eccv}

\usepackage{eccvabbrv}

\usepackage{comment}
\usepackage{graphicx}
\usepackage{booktabs}
\usepackage{circledsteps}
\usepackage{colortbl}
\usepackage{scalerel}
\usepackage{amsmath}
\usepackage{amssymb}
\usepackage{booktabs}
\usepackage{xcolor}
\usepackage{colortbl}
\usepackage{float}
\usepackage{hhline}
\usepackage{color, colortbl}
 \usepackage{soul}
 \usepackage{multirow}
\usepackage[tableposition=top]{caption}
\usepackage{fontawesome5}
\usepackage[accsupp]{axessibility}  % Improves PDF readability for those with disabilities.
\definecolor{lightgray2}{gray}{0.9}
\newcommand{\hlc}[2][yellow]{{
    \colorlet{foo}{#1}
    \sethlcolor{foo}\hl{#2}}
}
\usepackage{ulem}

\usepackage{hyperref}

\usepackage{orcidlink}

\begin{document}

\title{Synchronization is All You Need: Exocentric-to-Egocentric Transfer for Temporal Action Segmentation with Unlabeled Synchronized Video Pairs} 

\titlerunning{Synchronization is All You Need}

\author{Camillo Quattrocchi\inst{1}\orcidlink{0000-0002-4999-8698} \and
Antonino Furnari\inst{1,2}\orcidlink{0000-0001-6911-0302} \and
Daniele Di Mauro\inst{2}\orcidlink{0000-0002-4286-2050}  \and \\
Mario Valerio Giuffrida\inst{3}\orcidlink{0000-0002-5232-677X} \and
Giovanni Maria Farinella\inst{1,2}\orcidlink{0000-0002-6034-0432}}

\authorrunning{C.~Quattrocchi et al.}

\institute{Department of Mathematics and Computer Science, University of Catania, Italy \and
Next Vision s.r.l., Italy \and
School of Computer Science, University of Nottingham, United Kingdom}

\maketitle

\begin{abstract}
  We consider the problem of transferring a temporal action segmentation system initially designed for exocentric (fixed) cameras to an egocentric scenario, where wearable cameras capture video data. The conventional supervised approach requires the collection and labeling of a new set of egocentric videos to adapt the model, which is costly and time-consuming. Instead, we propose a novel methodology which performs the adaptation leveraging existing labeled exocentric videos and a new set of unlabeled, synchronized exocentric-egocentric video pairs, for which temporal action segmentation annotations do not need to be collected. We implement the proposed methodology with an approach based on knowledge distillation, which we investigate both at the feature and Temporal Action Segmentation model level. 
  %To evaluate our approach, we introduce a new benchmark based on the Assembly101 dataset.
  Experiments on Assembly101 and EgoExo4D demonstrate the  effectiveness of the proposed method against classic unsupervised domain adaptation and temporal alignment approaches. Without bells and whistles, our best model performs on par with supervised approaches trained on labeled egocentric data, without ever seeing a single egocentric label, achieving a $+15.99$ improvement in the edit score ($28.59$ vs $12.60$) on the Assembly101 dataset compared to a baseline model trained solely on exocentric data. In similar settings, our method also improves edit score by $+3.32$ on the challenging EgoExo4D benchmark. Code is available here: \url{https://github.com/fpv-iplab/synchronization-is-all-you-need}.
\end{abstract}

\section{Introduction}
\label{sec:intro}

\begin{figure}[h!]
  \centering
   \includegraphics[width=1\linewidth]{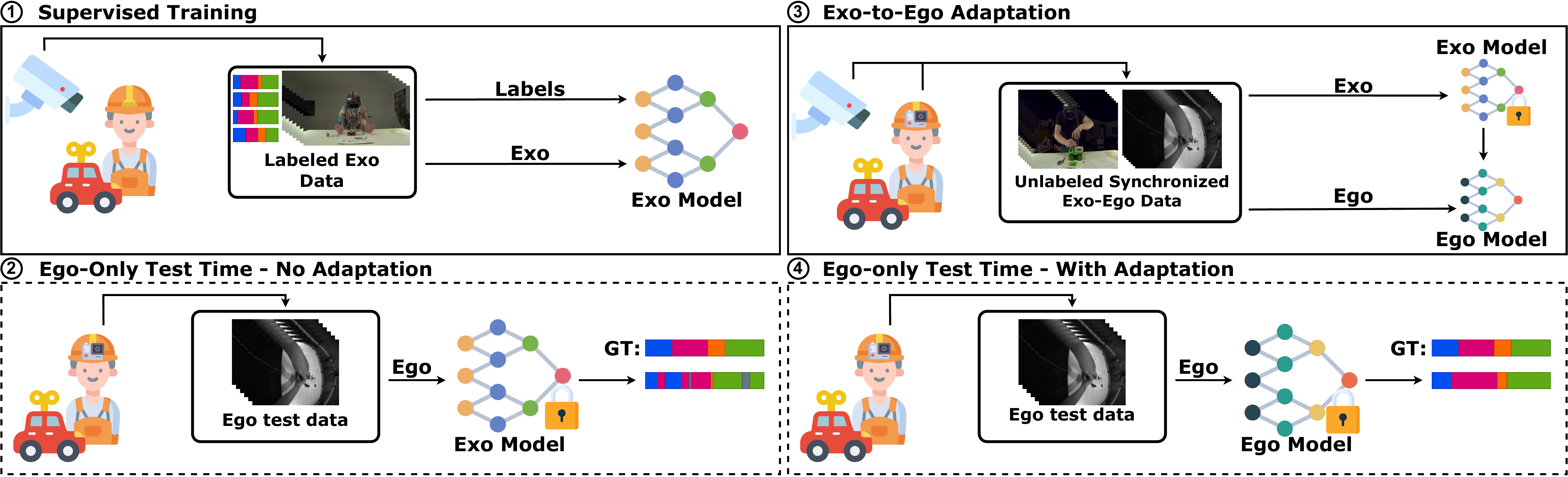} 
   \caption{
   We consider the problem of transferring a temporal action segmentation model from an exocentric to an egocentric setup.
   \Circled{1}~We assume that a set of labeled exocentric videos is available to train an exocentric temporal action segmentation model.
   \Circled{2}~When tested on egocentric data, the exocentric model exhibits poor performance due to domain shift.
   \Circled{3}~We propose to use an exocentric-to-egocentric adaptation process to adapt the exocentric model using a set of synchronized unlabeled exocentric-egocentric video pairs.
   \Circled{4}~At test time, the model is able to operate on egocentric data, with no access to exocentric videos.}
   \label{fig:problem}
\end{figure}

Temporal Action Segmentation (TAS) consists in dividing a video into meaningful temporal segments or intervals, where each segment corresponds to a distinct action or activity. 
TAS is important for different applications, ranging from collaborative robotics to the analysis of activities of daily living~\cite{lea2017temporal}.
TAS has been often studied in the context of \textit{exocentric} (or third-person) vision, where videos are collected through fixed cameras~\cite{8585084,singhania2022iterative,xu2022don,sayed2023new}. A line of work also considers the \textit{egocentric} (or first-person) domain, in which videos are collected by wearable devices~\cite{sener2022assembly101, 5204354, reza2023enhancing, sener2020temporal}.
Providing a privileged view on human-object interactions and object manipulation activities, the application of TAS to egocentric videos is relevant in different contexts ranging from work scenarios and assembly lines~\cite{chen2022long,sener2022assembly101} to understanding daily activities~\cite{li2018eye,Damen2020RESCALING}. Indeed, the exploitation of TAS in such contexts can be useful to analyze the user's behavior and provide assistive services through wearable devices.

Despite these potential benefits, current systems are mostly designed and tuned on exocentric videos collected and labeled by expert annotators. Due to the domain gap between exocentric and egocentric views, these systems tend to perform poorly when tested on videos collected through wearable cameras. 
TAS algorithms can be adapted to the new viewpoint through a supervised approach which consists in training or fine-tuning models on new egocentric videos of subjects performing the target activities, which have been purposely collected and labeled, a time-intensive and expensive process.
An alternative approach would consist in using unsupervised domain adaptation algorithms~\cite{wilson2020survey} to train the model using pre-existing labeled exocentric videos and newly collected unlabeled egocentric videos.
This approach is more convenient than the above-discussed standard supervised scheme as it only relies on unlabeled egocentric videos, thus avoiding the expensive annotation process when the model is adapted to the egocentric scenario.
However, as we also show in this paper, models trained with unsupervised domain adaptation tend to be affected by a residual domain shift that prevents full generalizability to the target domain (i.e., egocentric videos).

We observe that, in real application scenarios, it is inexpensive to collect an additional set of synchronized unlabeled \textit{exocentric-egocentric video pairs} of subjects performing the activities of interest.
For example, in a factory, workers can be equipped with wearable devices recording their activities while the already installed fixed cameras acquire synchronized videos of the same activities.\footnote{As we show in our experiments, our method works also when videos are not perfectly synchronized, hence sophisticated synchronization systems are not needed.}
The collected video pairs are naturally synchronized and they are not manually annotated for the temporal action segmentation task.
Therefore, we investigate whether such unlabelled synchronized video pairs can provide a suitable form of supervision to adapt an existing exocentric TAS model to the egocentric domain (see~\Cref{fig:problem}).
Based on the synchronized adaptation set, algorithms will aim to leverage domain-specific information learned from the source exocentric dataset to generate a model that performs well in the target egocentric video sequences. 
To tackle the proposed task, we introduce a novel exocentric-to-egocentric adaptation methodology based on knowledge distillation, investigating its application at the level of the feature extractor and downstream TAS model.
%To study the considered problem, we propose a new benchmark based on the Assembly101 dataset~\cite{sener2022assembly101}, which comprises multiple synchronized egocentric and exocentric video sequences. 
%Therefore, we introduce an exocentric-to-egocentric adaptation methodology based on knowledge distillation, investigating its application at the level of the feature extractor and downstream TAS model.
We perform experiments on two different view adaptation settings of the Assembly101~\cite{sener2022assembly101}, which comprises multiple synchronized egocentric and exocentric video sequences of subjects assembling toys, and on the recently released EgoExo4D dataset~\cite{grauman2023ego}, which includes multi-view synchronized sequences of subjects performing diverse activities.
Experiments confirm that synchronized video pairs provide a valuable form of supervision, with our best model bringing improvements of $+15.99$ and $+11.64$ edit score on Assembly101 in the two view transfer settings, compared to baselines trained solely on exocentric data, and recovering the exo-ego domain gap in EgoExo4D, with improvements of up to $+3.32$ in edit score.

Our contributions are as follows: 1) we propose and investigate a new exocentric-to-egocentric adaptation task which aims to transfer an already trained exocentric TAS system to egocentric cameras in the presence of cross-view synchronized video pairs; 2) we design and benchmark an adaptation methodology which performs knowledge distillation at the level of the feature extractor and downstream TAS model; 3) we benchmark the proposed task on the Assembly101~\cite{sener2022assembly101} and EgoExo4D~\cite{grauman2023ego} datasets, reporting results on different view adaptation settings.
Experiments confirm the effectiveness of the proposed methodology with significant improvements over current approaches, suggesting that synchronization allows to recover the performance of supervised methods, without relying on egocentric labels. 
%We will release the code of our methods, the dataset splits, and the pre-extracted features to enable future research.
%The code of our methods and the dataset splits are available at the following URL: anonymous for double blind submission.% \url{https://github.com/fpv-iplab/synchronization-is-all-you-need}.
Code is available here: \url{https://github.com/fpv-iplab/synchronization-is-all-you-need}.

\section{Related Work}
\label{sec:related_work}

\noindent
\textbf{Temporal Action Segmentation}
The literature on Temporal Action Segmentation is rich, with many approaches based on classic~\cite{5204354} or modern~\cite{liu2023diffusion} techniques.
Among recent representative methods, the authors of~\cite{lea2017temporal} first proposed an encoder-decoder model based on temporal convolutional networks.
The authors of~\cite{li2020ms} later proposed a multistage architecture based on temporal convolutional networks that refines the predicted actions in multiple stages. In~\cite{8585084}, it was presented an approach based on a recurrent neural network that models combinations of discriminative representations of subactions, allowing temporal alignment and inference over long sequences. 
The authors of~\cite{singhania2021coarse} proposed C2F-TCN, a temporal encoder-decoder architecture based on an implicit ensemble of multiple temporal resolutions to mitigate sequence fragmentation.
The study in~\cite{singhania2022iterative}~showed that it is possible to learn higher-level representations to interpret long video sequences. 
The authors of~\cite{xu2022don} proposed DTL, a framework that uses temporal logic to constrain the training of action analysis models improving the performance of different architectures.
The authors of~\cite{liu2023diffusion} use denoising diffusion models to iteratively refine predicted actions starting from random noise, using video features as conditions.
The reader is referred to~\cite{ding2022temporal} for a recent survey on the topic of Temporal Action Segmentation.
We follow~\cite{sener2022assembly101} and consider a Coarse to Fine Temporal Convolutional Network (C2F-TCN)~\cite{singhania2021coarse} as a temporal action segmentation approach.
%Temporal Shift Module (TSM)~\cite{lin2019tsm} for feature extractor and Coarse to Fine Temporal Convolutional Network (C2F-TCN)~\cite{singhania2021coarse} as a temporal action segmentation approach. A DINOv2~\cite{oquab2023dinov2} feature extractor is also considered in our study.\\
\\
\noindent
\textbf{Exocentric-to-Egocentric Model Transfer}
Few previous works considered the problem of transferring knowledge from exocentric to egocentric views. 
The authors of~\cite{sigurdsson2018actor} proposed the Charades-Ego dataset containing pairs of first- and third-person synchronized videos, labeled for action recognition. A model jointly learning to represent those two domains with a triplet loss was also proposed.
The approach presented in~\cite{yu2019see} learns a shared representation of exocentric and egocentric videos by modeling joint attention maps to link the two viewpoints. 
The authors of~\cite{Li_2021_CVPR} proposed to leverage latent signals in third-person videos that are predictive of egocentric-specific properties, such as camera position, manipulated objects, and interactions with the environment to kickstart the training of an egocentric video model on exocentric data.
In~\cite{xue2023learning}, a method to jointly learning from unpaired egocentric and exocentric videos was proposed, with the aim to leverage temporal alignment as a self-supervised learning objective and learn view-invariant features. 
The work in~\cite{li2023locate} introduced a method to extrapolate affordance grounding from image-level labels on exocentric images and transfer them to egocentric views of the same objects.
Previous approaches focused on exocentric-to-egocentric transfer in the absence of synchronized video pairs and considered tasks other than Temporal Action Segmentation. Ours is the first systematic investigation on unlabeled synchronized exo-ego video pairs as self-supervision for exo-to-ego model transfer for Temporal Action Segmentation.\\
%\noindent
\textbf{Domain Adaptation}
Domain Adaptation methods aim to leverage the knowledge gained from a specific source domain to improve performance of a machine learning model on a target domain in the presence of a distribution shift~\cite{DBLP:journals/corr/Csurka17}.
In the unsupervised domain adaptation setting, models assume the presence of source labeled data and target unlabeled data during training.
In recent years, many domain adaptation techniques have been proposed for several computer vision tasks, including some related to video understanding~\cite{choi2020shuffle, kim2021learning}. 
Different works focused on the adaptation of action recognition models when source and target domains are given by different environments. 
Among these works, the authors of~\cite{Chen_2019_ICCV} introduced two large-scale datasets for video domain adaptation for action recognition, \textit{UCF-HMDB$_{full}$} and \textit{Kinetics-Gameplay}, comprising both real and virtual domains, and proposed the Temporal Attentive Adversarial Adaptation Network (TA3N) that simultaneously attends, aligns, and learns temporal dynamics across domains, achieving state-of-the-art performance. 
In~\cite{munro2020multi}, the RGB and optical flow modalities are leveraged to reduce the domain shift arising from training and testing action recognition models in different environments. The authors of~\cite{Kim_2021_ICCV} investigated video domain adaptation for action recognition in a cross-modal contrastive learning framework. The authors of~\cite{10139790} introduced \textit{ADL-7} and \textit{GTEA-KITCHEN-6}, two egocentric domain-adaptation datasets for action recognition, and proposed a Channel Temporal Attention Network (CTAN) architecture.
The work of~\cite{choi2020unsupervised} demonstrated promising performance on the task of adapting first- and third-person videos for drone action recognition.
The problem of domain adaptation for temporal action segmentation was investigated in~\cite{chen2020action}, where different domains arise from different environments.
Previous works mainly focused on action recognition~\cite{Chen_2019_ICCV,munro2020multi,choi2020unsupervised} and considered different environments or subjects as a form of distribution shift~\cite{Chen_2019_ICCV,munro2020multi,10139790} with few works tackling exocentric-to-egocentric adaptation as a domain adaptation problem~\cite{choi2020unsupervised} or considering temporal action segmentation as a target task~\cite{chen2020action}.
Also, previous works considered unpaired inputs from the different domains, which limit the amount of supervision that can be leveraged for the model transfer.
Our investigation is first to consider exo-to-ego adaptation for temporal action segmentation in the presence of synchronized unlabeled video pairs, which, as we show in the experiments, provide a strong form of self-supervision for adaptation.\\
%\noindent 
\textbf{Knowledge Distillation}
Early knowledge distillation methods were designed to transfer knowledge from a resource intensive neural network (the ``teacher'') to a less computationally demanding network (the ``student'')~\cite{hinton2015distilling}.
These techniques exploited the teacher model's logits to supervise the student network through a Kullback–Leibler divergence loss on the probability distributions predicted by the two models.
An alternate set of strategies proposed to align the intermediary activations of teacher and student networks to provide a more accurate distillation signal~\cite{Huang2017,romero2014fitnets}.
These approaches stem from the observation that neural networks internally develop hierarchical representations of their inputs, which characterize the way final predictions are made.
Another line of approaches proposed to model the relationships between activations in consecutive layers of the teacher network as a way of guiding the student's learning~\cite{yim2017gift,passalis2020heterogeneous}. 
Beyond adapting a larger model to a smaller one, knowledge distillation has also been used to improve the results of action recognition~\cite{crasto2019mars,liu2019attention,stroud2020d3d}, early action prediction~\cite{wang2019progressive}, and action anticipation~\cite{camporese2021knowledge,fernando2021anticipating,tran2019back,furnari2023streaming}.
Unlike the aforementioned works, we use knowledge distillation to adapt a model trained solely on exocentric videos to generalize over egocentric videos based on a set of unlabeled synchronized exocentric-egocentric video pairs. 
We investigate distillation at the levels of feature extraction and TAS model, showing that unlabeled synchronized exo-ego pairs provide a strong learning objective for exo-to-ego model transfer.

\section{Methodology}
\label{sec:methodology}

%\subsection{Problem Definition}
\textbf{Problem Definition}
Let $\mathcal{D}_{train}^{exo} = \{(V_i^{exo}, S_i^{exo})\}_{i=1}^n$ be a collection of $n$ labeled exocentric videos, where $V_i^{exo}$ is the $i^{th}$ exocentric video of the collection and $S_i^{exo}$ is the corresponding manually labeled segmentation. Let $\mathcal{D}_{adapt}^{pair} = \{(V_j^{exo}, V_j^{ego})\}_{j=1}^m$ be a collection of $m$ time-synchronized unlabeled exocentric-egocentric video pairs, designed to perform the adaptation process. 
Note that these video pairs can be collected with existing egocentric and exocentric systems and do not need to be labeled.
We define exocentric-to-egocentric domain adaptation as the problem of training a TAS algorithm $\Psi$ on both $\mathcal{D}_{train}^{exo}$ and $\mathcal{D}_{adapt}^{pair}$ to generalize over a test set $\mathcal{D}_{test}^{ego}=\{(V_h^{ego},S_h^{ego})\}_{h=1}^{q}$ of egocentric videos $V_h^{ego}$ and the corresponding ground truth temporal segmentations $S_h^{ego}$ (used only for evaluation purposes). 
The proposed task is related to two other problems previously investigated in the literature: 1) standard supervised TAS, that directly trains on a labeled dataset of egocentric videos $\mathcal{D}_{train}^{ego}$~\cite{ding2022temporal}; 2) cross-view domain adaptation~\cite{Li_2021_CVPR}, which trains on a labeled set of exocentric videos $\mathcal{D}_{train}^{exo}$ and an unlabeled set of egocentric videos $\mathcal{D}_{adapt}^{ego}=\{V_j^{ego}\}_{j=1}^m$ for adaptation. Differently from 1), we aim to generalize to egocentric videos using only labeled exocentric videos of $\mathcal{D}_{train}^{exo}$ and a set $\mathcal{D}_{adapt}^{pair}$ 
 of unlabeled exocentric-egocentric synchronized video pair, whereas, differently from 2), we make use of a set of exocentric-egocentric video pairs $\mathcal{D}_{adapt}^{pair}$, which are deemed to provide a stronger supervision than the set of egocentric videos~$\mathcal{D}_{adapt}^{ego}$.
\subsection{Proposed Adaptation Approach}
\label{sec:compared_methods}
\begin{figure}[tp]
\centering
\includegraphics[width=0.8\linewidth]{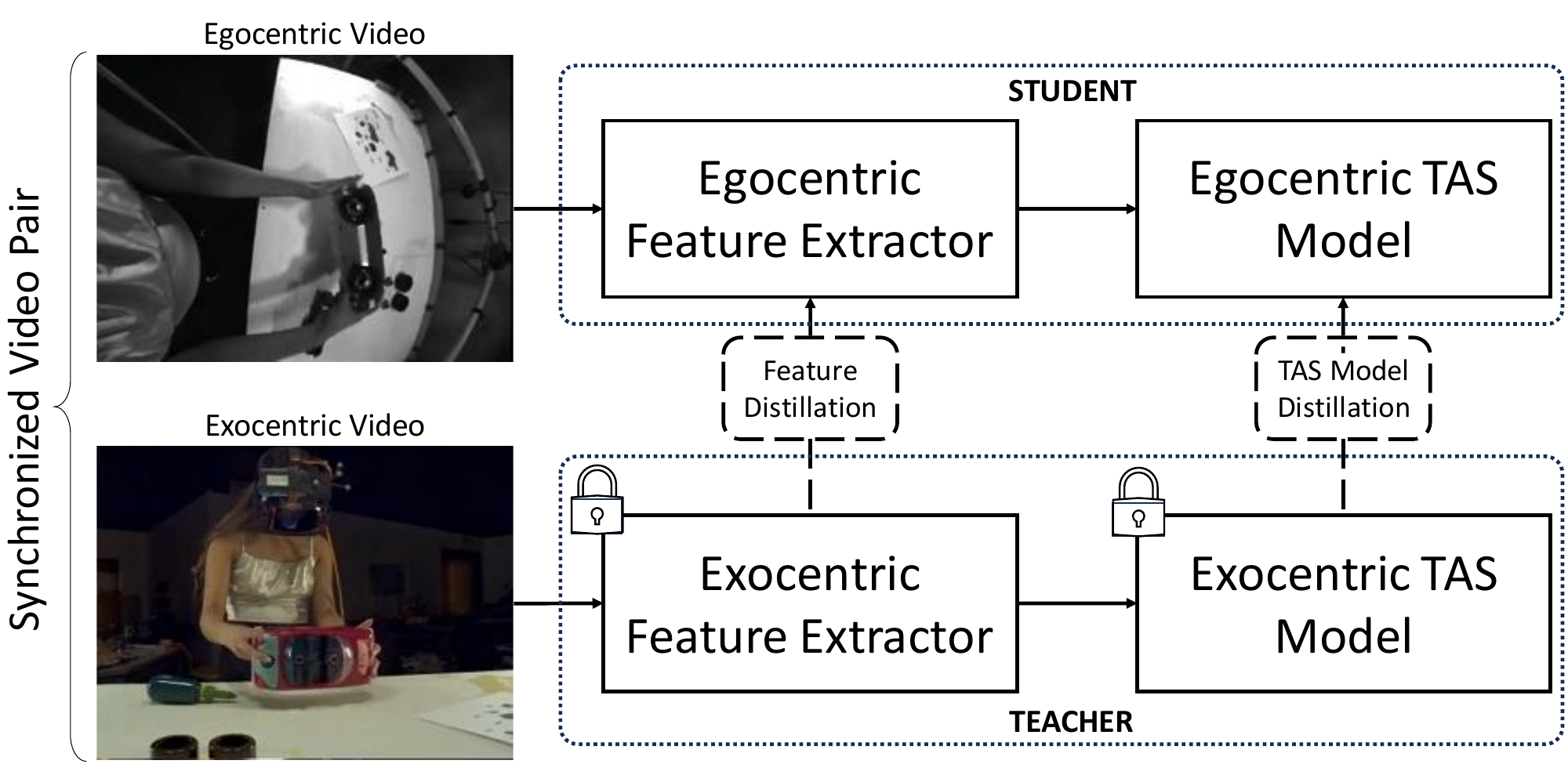}
    \caption{The proposed method to adapt an exocentric Temporal Action Segmentation model trained on $\mathcal{D}_{train}^{exo}$ to an egocentric setting using a set of unlabelled synchronized video pairs $\mathcal{D}_{adapt}^{pair}$. We investigate distillation at two different levels: the feature extractor and the model. The bottom branch (exocentric) is used to supervise the top branch (egocentric).}
    \label{fig:flow}
\end{figure}

We propose a methodology to adapt an exocentric TAS model to egocentric videos that relies on the distillation of knowledge at two different levels: the feature extractor and the downstream TAS model. See Figure~\ref{fig:flow}.
%illustrates the proposed methodology. 

\noindent
\textbf{Algorithm Factorization} Our model $\Psi$ is factorized into $\Psi = \gamma \circ \phi$, where $\phi$ is a feature extractor and $\gamma$ is the TAS model making predictions based on the features extracted by $\phi$. 
%We experiment with two feature extractors $\phi$: a domain-specific TSM~\cite{lin2019tsm} model trained to perform action recognition, as adopted in~\cite{sener2022assembly101}, and an off-the-shelf domain-agnostic DINOv2 model~\cite{oquab2023dinov2}.
%We adopt C2F-TCN~\cite{singhania2021coarse} as the prediction algorithm $\gamma$.

\noindent
\textbf{Teacher Model} We first train a teacher model by optimizing the feature extractor $\phi^T$ and the downstream TAS algorithm $\gamma^T$ on the exocentric supervised training set $\mathcal{D}_{train}^{exo}$. 
We train the C2F-TCN model $\gamma^T$ on top of features extracted by the feature extractor $\phi^T$. After training, the parameters of the teacher model are fixed. The teacher is no longer optimized during the distillation process.

\noindent
\textbf{Student Model} The student model shares the same architecture as the teacher model, adopting the same feature extractor $\phi^S$ and prediction algorithm $\gamma^S$. 

\noindent
\textbf{Distillation Process} During the distillation process, the student is optimized to mimic the outputs of the teacher component. This process is carried out using the adaptation set $\mathcal{D}_{adapt}^{pair}$. No exocentric or egocentric labels are used at this time, with the only supervision being given by the exocentric videos passed through the fixed teacher. We consider two adaptation steps that can be either included separately or combined: feature distillation and TAS model distillation. 

\textit{Feature Distillation} In the feature distillation process, we sample pairs of video segments $(S_j^{exo}, S_j^{ego})$ from video pairs $(V_j^{exo}, V_j^{ego}) \in \mathcal{D}_{adapt}^{pair}$ and train $\phi^S$ to minimize the following Mean Squared Error (MSE) loss function:

\begin{equation}
\mathcal{L}_f=\frac{1}{M}\sum_{j=1}^{M}||\phi^S(S_j^{ego}) - \phi^T(S_j^{exo})||^2    
\end{equation}

\noindent
where $M$ is the batch size. In practice, we encourage the student model to extract from the egocentric videos representations similar to the exocentric ones extracted by the teacher model. 
%We apply this procedure to the TSM model. Since fine-tuning the DINOv2 feature extractor is highly demanding in terms of computation, we instead train an adapter module $\theta$ with a residual connection which takes as input DINOv2 features extracted by the student model and learns to adapt them with the same MSE loss. We implement $\theta$ as a transformer encoder network that translates video feature sequences in one pass. The transformer network has $12$ layers, $16$ heads, and a number of hidden units equal to $1024$. At test time, the adapted DINOv2 features are extracted by the composed function $\theta(\phi(\cdot))$.
Additional information about the Feature Distillation process is available in supplementary material.

\textit{TAS Model Distillation} This process aims to distill the student TAS model $\gamma^S$ to make its internal representations similar to those extracted by $\gamma^T$ on the paired exocentric video. Formally, let $\gamma_{l}(V_j)$ be the features extracted by the prediction model $\gamma$ at layer $l$ of the architecture when observing egocentric video. We distill the student module by minimizing the following MSE loss function:

\begin{equation}
\mathcal{L}_m = \frac{1}{M}\sum_{j=1}^M \sum_{l=1}^L ||\gamma^S_l(\phi^S(V_j^{ego})) - \gamma^T_l(\phi^T(V_j^{exo}))||^2
\end{equation}

\noindent
where $\phi(V)$ represents a sequence of frame-wise features densely extracted from video $V$ using the feature extractor $\phi$, $M$ is the batch size, $L$ is the number of layers of the model $\gamma$ and $(V_j^{ego}, V_j^{exo}) \in \mathcal{D}_{adapt}^{pair}$.
Additional information about the TAS Model Distillation process is available in supplementary material.
\begin{figure}[tp]
\centering
    \includegraphics[width=0.328\textwidth]{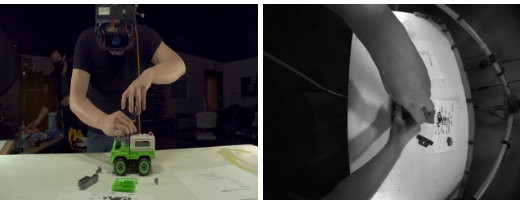}
    \hfill
    \includegraphics[width=0.328\textwidth]{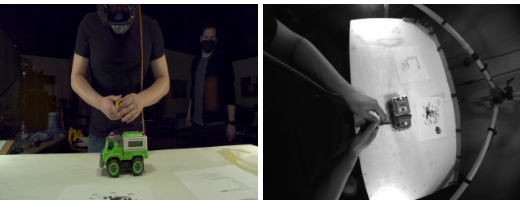}
    \hfill
    \includegraphics[width=0.328\textwidth]{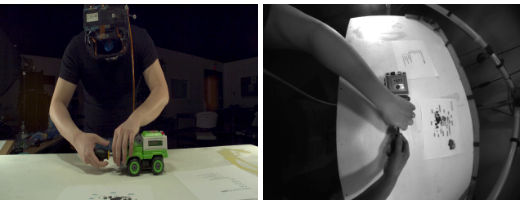}
   \caption{Three pairs of synchronized exo-ego (v3, e4) frames from Assembly101.}
   \label{fig:views}
\end{figure} 
\section{Experimental Settings}
\label{sec:experimental_setup}
%\subsection{Datasets and Implementation Details}
%\noindent
\textbf{Assembly101 Dataset}
We base our main experiments on the Assembly101 dataset~\cite{sener2022assembly101}.
The dataset contains 4321 videos of subjects performing assembly and disassembly procedures of toy models.
Videos are collected from multiple synchronized views:
4 first-person monochrome cameras mounted on a custom headset with a resolution of $640\times480$ pixels, and 8 third-person RGB cameras (5 mounted overhead, 3 on the side) with a resolution of $1920\times 1080$ pixels. 
The videos contain frame-wise annotations of fine-grained and coarse-grained action annotations. 
As in~\cite{sener2022assembly101}, we consider coarse-grained action annotations for the temporal action segmentation experiments.
For our experiments, we selected two exocentric views (\textit{v3} and \textit{v4}) and one egocentric view (\textit{e4}). This allows us to consider two different adaptation settings $v3 \to e4$ and $v4 \to e4$, which allow us to assess the generalization of our method.\footnote{More information on view selection in supplementary material.}
Each view is associated to $362$ videos, which we split as described in the following.
%, which we split in a $\sim 40:20:40$ ratio\footnote{We choose this ratio to provide enough training and test data, while still relying on a small set of synchronized unlabeled pairs.} as detailed in the following. 
We randomly selected $143$ exocentric videos to form the exocentric training set $\mathcal{D}_{train}^{exo}$. The corresponding synchronized egocentric videos are included in a $\mathcal{D}_{train}^{ego}$ set.
A set of $63$ exocentric-egocentric synchronized video pairs is selected to form the adaptation set $\mathcal{D}_{adapt}^{pair}$, whereas $156$ egocentric videos are selected to form the egocentric test set $\mathcal{D}_{test}^{ego}$. The corresponding exocentric videos are included in a set $\mathcal{D}_{test}^{exo}$.
Examples of exocentric-egocentric video pairs from the dataset are reported in~\Cref{fig:views}.
Following the authors~\cite{sener2022assembly101}, we use TSM~\cite{lin2019tsm} as the feature extractor $\phi$ and C2F-TCN~\cite{singhania2021coarse} as the TAS model. We also experiment with DINOv2~\cite{oquab2023dinov2} features densely extracted over all video frames.
Since fine-tuning DINOv2 is highly demanding in terms of computation, for feature distillation we train an adapter module $\theta$ with a residual connection which learns to adapt DINOv2 features extracted by the student model with an MSE loss and use $\theta(\phi(\cdot))$ as a feature extractor at test time.
%We implement $\theta$ as a transformer encoder network that translates video feature sequences in one pass. The transformer network has $12$ layers, $16$ heads, and a number of hidden units equal to $1024$. At test time, the adapted DINOv2 features are extracted by the composed function $\theta(\phi(\cdot))$.
%Teacher and student TSM models are pre-trained on EPIC-KITCHENS-100~\cite{Damen2020RESCALING} as specified in~\cite{sener2022assembly101}. We use the code provided by the authors of~\cite{sener2022assembly101} to train the TSM and C2F-TCN models. We use the \textit{large} DINOv2 model to extract per-frame features using the official implementation\footnote{\url{https://github.com/facebookresearch/dinov2}}. Additional implementation details are in Appendix~\ref{supp:training_details} of the supp. material.
We will release all the details of the proposed split and extracted features to support future research.\footnote{Additional implementation details are in supplementary material.}
%All the details of the proposed split and the extracted features to support future research on the proposed task are available at the following URL: anonymous for double blind submission. %\url{https://github.com/fpv-iplab/synchronization-is-all-you-need}.

\noindent
\textbf{EgoExo4D Dataset}
We also consider the recently proposed EgoExo4D~\cite{grauman2023ego} dataset, which contains multi-view videos of subjects performing different activities in diverse scenarios, totaling 1,422 hours of footage.
Each video is acquired from an egocentric device and different synchronized GoPro cameras.
For each video, we consider the egocentric view and the exocentric one marked as \textit{best view} by humans annotators.
We considered the subset of the dataset containing procedural activities, where each video includes hierarchical keystep annotations breaking down human activities in temporal video segments.
Similarly to Assembly101, we use parent annotations, which are akin to coarse labels.
%a multimodal and multiview dataset focusing on capturing synchronized first and third person views. It showcases proficient human actions like sports, cooking, music, dance, and bike repair. Gathered by a global cohort of over 800 individuals across 13 cities, these activities unfolded within 131 distinct natural settings, yielding lengthy video recordings ranging from 1 to 42 minutes each, totaling 1,422 hours of footage. Each video is complemented by multichannel audio, eye gaze tracking, 3D point clouds, camera positions, IMU data, and multiple corresponding textual descriptions. %The dataset introduces a range of benchmark tasks and their annotations, including fine-grained activity understanding, proficiency estimation, cross-view translation, and 3D hand/body pose.
It is worth noting that this dataset provides a much more challenging benchmark as compared to Assembly101, due to the non-standardized placement of cameras, whose placement can vary from a video-to-video basis.
To reduce complexity, we selected 5 of the most representative scenarios in the subset of cooking activities: \textit{Making Cucumber and Tomato Salad}, \textit{Cooking Scrambled Eggs}, \textit{Cooking an Omelet}, \textit{Making Sesame-Ginger Asian Salad}, \textit{Cooking Tomato and Eggs}.
This leads to $233$ videos, which we randomly split in $90$ exocentric videos for exocentric training ($\mathcal{D}_{train}^{exo}$), the associated $90$ egocentric videos for egocentric training ($\mathcal{D}_{train}^{ego}$), $89$ exo-ego video pairs for adaptation ($\mathcal{D}_{adapt}^{pair}$), and $44$ exocentric and paired egocentric videos for test ($\mathcal{D}_{test}^{exo}$ and $\mathcal{D}_{test}^{ego}$).
We obtain a set of $32$ coarse action classes plus a \textit{Background} class, arising from unlabeled parts of the video due to the fact that annotations were not originally designed for temporal action segmentation. To conform to the standard temporal action segmentation scenario in which the background class is generally not present~\cite{sener2022assembly101}, we do not consider this class during training and evaluation.
We use C2F-TCN for TAS as in~\cite{sener2022assembly101} and adopt Omnivore~\cite{girdhar2022omnivore} features provided by the authors.\footnote{\url{https://ego-exo4d-data.org/}}

\noindent
\textbf{Evaluation Measures}
All models are evaluated using standard measures, as reported in previous TAS works~\cite{sener2022assembly101,lea2017temporal,ding2022temporal}. In particular, we report two segment-based evaluation measures, namely edit score and F1 score, and a frame-based evaluation measure, namely Mean over Frames (MoF).
All results in this paper are reported in percentage ($0-100$ range). See~\cite{ding2022temporal} for more information on the definition of the evaluation measures.

\section{Results and Discussion}
%In the following sections, we first show our main analysis on Assembly101, then show the generalization of our approach to the challenging scenario of EgoExo4D.

\subsection{Performance of the Proposed Approach}
\Cref{tab:main_results} compares different approaches to exo-to-ego adaptation on the $v3 \to e4$ view adaptation setting. Exo-oracle baselines (lines 1-2) are trained and tested on exocentric data. These are our teacher models, and are reported here for reference. Ego-oracle baselines (lines 3-4) are trained and tested on egocentric data. They represent the typical results of a supervised model making use of labeled egocentric data. Lines 5-14 report the results of different variations of the proposed method based on knowledge distillation. We compare different choices of backbones and distillation modules.

\noindent
\textbf{Oracle Performance}
We note that exo-oracle methods (lines 1-2) systematically outperform ego-oracle approaches (lines 3-4) according to all evaluation measures when the same backbone is considered (e.g., $31.45$ vs $29.25$ edit score and $29.92$ vs $25.40$ $F1_{25}$ comparing line 1 with 3). Note that these approaches have the same architecture but take as input different videos of the same scene (i.e., the egocentric and exocentric views). Hence, the observed difference in performance is likely due to the additional challenges involved with the processing of egocentric video in the Assembly101 dataset, including occlusions, motion blur, and the use of grayscale images (see Figure~\ref{fig:views}).
In this context, the domain-specific TSM model achieves better performance than the domain-agnostic DINOv2 model (e.g., $31.45$ vs $28.55$ edit score and $20.88$ vs $17.79$ $F1_{50}$ comparing line 1 with 2 and $29.25$ vs $26.42$ edit score comparing line 3 with 4).
However, remarkably, the gap between DINOv2 and TSM features is small despite DINOv2 has not been trained on any domain-specific videos (e.g., $-2.9$ edit score comparing line 2 with 1 and $-2.83$ when comparing line 4 with 3). This highlights the potential of DINOv2 features as a suitable off-the-shelf feature extractor in the considered setup.
\begin{table*}[t]
    \begin{minipage}[t]{\textwidth}
    \centering
        \caption{Comparison of different approaches on the v3 $\to$ e4 view adaptation setting. Both feature and TAS model distillation allow to reduce the gap between the baselines (highlighted). Best results within groups are reported in bold. Best results among the adaptation methods are underlined.}
        \vspace{1mm}
    \resizebox{\linewidth}{!}{
    \begin{tabular}{r|c|c|cc|rrrrr}
    \multicolumn{1}{c}{} & \multicolumn{1}{c}{\textbf{Adaptation Task}} & \multicolumn{1}{c}{\textbf{Backbone}} &  \multicolumn{1}{c}{\textbf{Feature Dist.}} & \multicolumn{1}{c}{\textbf{TAS Dist.}} & \multicolumn{1}{c}{\textbf{Edit}} & \multicolumn{1}{c}{$\bf{F1_{10}}$} & \multicolumn{1}{c}{$\bf{F1_{25}}$} & \multicolumn{1}{c}{$\bf{F1_{50}}$} & \multicolumn{1}{c}{\textbf{MoF}} \\
    \cline{1-10}
    1&\multirow{2}{*}{\makecell{EXO $\to$ EXO\\(exo-oracle)}} & TSM & & & \textbf{31.45} & \textbf{34.24} & \textbf{29.92} & \textbf{20.88} & \textbf{37.39} \\
    2&& DINOv2 & & & 28.55 & 30.13 & 25.84 & 17.79 & 36.03 \\
    \cline{1-10}
    3&\multirow{2}{*}{\makecell{EGO $\to$ EGO\\(ego-oracle)}} & TSM & & & \textbf{29.25} & \textbf{31.06} & \textbf{25.40} & \textbf{17.47} & \textbf{36.60}   \\
    4&& DINOv2 & & & 26.42 & 26.50 & 22.16 & 14.84 & 33.47   \\
    \hline
    5&\multirow{9}{*}{EXO $\to$ EGO}&\multirow{4}{*}{TSM} & & & \cellcolor{lightgray2}10.53 & \cellcolor{lightgray2}6.47 & \cellcolor{lightgray2}2.44 & \cellcolor{lightgray2}0.56 & \cellcolor{lightgray2}5.42 \\
    6&& & \checkmark & & 12.38 & 8.73 & 5.64 & 1.99 & 8.21 \\
    7&& &  & \checkmark & 20.17 &	21.11 &	18.23 &	13.12 &	17.77 \\
    8&& & \checkmark & \checkmark &  \textbf{25.39}	&\textbf{26.79}	&\textbf{22.37}	&\textbf{15.52}	&\textbf{30.30}    \\
    \cline{3-10}
    9&&\multicolumn{3}{l|}{Improvement w.r.t. \hlc[lightgray2]{baseline (line 5)}} & $+14.86$ & $+20.32$ & $+19.93$ & $+14.96$ & $+24.88$ \\
    \hhline{~|~|-|--|-----}
    10&&\multirow{4}{*}{DINOv2} & & & \cellcolor{lightgray2}12.60 & \cellcolor{lightgray2}10.18 & \cellcolor{lightgray2}7.21 & \cellcolor{lightgray2}2.40 & \cellcolor{lightgray2}14.15 \\
    11&& & \checkmark & & 14.38 & 10.80 & 6.53 &	2.33 &	10.94  \\
    12&& & & \checkmark & \textbf{\underline{28.59}} & \textbf{\underline{29.58}} & \textbf{\underline{24.84}} &	\textbf{\underline{16.38}} &	31.36  \\
    13&& & \checkmark & \checkmark & 28.13 &	28.75 &	24.16 &	15.79 &	\underline{\textbf{32.67}}    \\
    \cline{3-10}
    14&&\multicolumn{3}{l|}{Improvement w.r.t. \hlc[lightgray2]{baseline (line 10)}} & $+15.99$ & $+19.40$ & $+17.63$ & $+13.98$ & $+18.52$ \\
    \cline{1-10}
    
    \end{tabular}
    }
\label{tab:main_results}
\hfill
\end{minipage}
\begin{minipage}[t]{\textwidth}
\centering
        \caption{Results for the v4 $\to$ e4 view adaptation setting. Best results within groups are reported in bold. Best results among the adaptation methods are underlined.}
        \vspace{1mm}
    \resizebox{\linewidth}{!}{
    \begin{tabular}{r|c|c|cc|rrrrr}
    \multicolumn{1}{c}{} & \multicolumn{1}{c}{\textbf{Adaptation Task}} & \multicolumn{1}{c}{\textbf{Backbone}} &  \multicolumn{1}{c}{\textbf{Feature Dist.}} & \multicolumn{1}{c}{\textbf{TAS Dist.}} & \multicolumn{1}{c}{\textbf{Edit}} & \multicolumn{1}{c}{$\bf{F1_{10}}$} & \multicolumn{1}{c}{$\bf{F1_{25}}$} & \multicolumn{1}{c}{$\bf{F1_{50}}$} & \multicolumn{1}{c}{\textbf{MoF}} \\
    \cline{1-10}
    \hphantom{1}1&EXO $\to$ EXO& DINOv2 & & & 24.03 & 19.78 & 15.17 & 8.17 & 22.54 \\
    \cline{1-10}
    2&EGO $\to$ EGO& DINOv2 & & & 26.42 & 26.50 & 22.16 & 14.84 & 33.47   \\
    \hline
    3&\multirow{5}{*}{EXO $\to$ EGO}&\multirow{4}{*}{DINOv2} & & & \cellcolor{lightgray2}12.82 & \cellcolor{lightgray2}10.88 & \cellcolor{lightgray2}6.93 & \cellcolor{lightgray2}2.62 & \cellcolor{lightgray2}10.65 \\
    4&& & \checkmark & & 20.24 & 21.16 & 17.90 & 9.90 &	25.18  \\
    5&& & & \checkmark & 24.23 &  22.93 &  18.21 &	11.91 & 25.87  \\
    6&& & \checkmark & \checkmark & \textbf{\underline{24.46}} &	\textbf{\underline{25.80}} & \textbf{\underline{21.32}} & \textbf{\underline{13.87}} & \textbf{\underline{29.39}}    \\
    \cline{3-10}
    7&&\multicolumn{3}{l|}{Improvement w.r.t. \hlc[lightgray2]{baseline (line 3)}} & $+11.64$ & $+14.92$ & $+14.39$ & $+11.25$ & $+18.74$ \\
    \cline{1-10}
    
    \end{tabular}}
\label{tab:view}  
\hfill
\end{minipage}

\end{table*}
\\
\noindent
\textbf{Contributions of the Adaptation Modules}
The proposed adaptation scheme based on knowledge distillation achieves remarkable results both when using domain-specific TSM features (lines 5-9) and when using domain-agnostic DINOv2 features (lines 10-14).
For TSM features, we observe consistent improvements when the feature distillation module is activated (comparing lines 6-5, we obtain $12.38$ vs $10.53$ edit score and $8.21$ vs $5.42$ MoF). Larger improvements are obtained with TAS model distillation (comparing 7 vs 5-6, we obtain an edit score of $20.17$ vs $10.53$ and $12.38$). Best results are obtained when both feature distillation and TAS model distillation are activated, with significant $+24.88$, $+20.32$ and $+14.86$ improvements in MoF, $F1_{10}$ and edit score (see line 9).
\\
Similar trends are observed with DINOv2 features, with feature distillation and TAS model distillation bringing improvements with respect to the baseline (compare lines 12-13 vs line 10).
Differently from results based on TSM, best overall results are already achieved activating the TAS model distillation, with consistent improvements of $+19.40$, $+17.21$ and $+15.99$ in $F1_{10}$, $MoF$ and edit score respectively (see line 14). Combining both feature and TAS model distillation leads to comparable performance on all evaluation measures, with the exception of MoF which its improved from $31.36$ to $32.67$ (compare lines 12-13). This marginal improvement suggests the effectiveness of DINOv2 features, which being generic, do not require an explicit adaptation step.
Comparing line 10 with line 5, we can observe that the domain-agnostic DINOv2 feature extractor exhibits a smaller exo-ego domain gap (e.g., $12.60$ vs $10.53$ edit score and $14.15$ vs $5.42$ MoF). This smaller gap also affects best overall results (numbers in bold), with the best model based on DINOv2 (line 12) obtaining higher performance measures when compared with the best model based on TSM (line 8), e.g., $28.59$ vs $25.39$ edit score, and $24.84$ vs $22.37$ $F1_{25}$. 
\\
\noindent
\textbf{Adaptation Approaches versus Oracle Baselines}
Interestingly, our best results based on DINOv2 closely match the results obtained by the best ego-oracle baseline (line 12 vs line 3 - e.g., $28.59$ vs $29.25$ edit score, and $29.58$ vs $30.13$ $F1_{10}$) and outperform the results obtained by the ego-oracle baseline based on the same DINOv2 feature extractor (line 12 vs line 4 - e.g., $28.59$ vs $26.42$ edit score, and $24.84$ vs $22.16$ $F1_{25}$). This highlights that, in the presence of labeled exocentric videos, unlabeled synchronized exo-ego video pairs, and an effective adaptation technique, labeling egocentric videos could be entirely avoided. 
\\
%\noindent
\textbf{Multiple View Transfer Settings}
Table~\ref{tab:view} reports the results on the second view adaptation setting $v4 \to e4$. Note that, being the target view $e4$ the same as Table~\ref{tab:main_results}, results in the two tables are comparable.
In this table, we report results only with DINOv2, which achieved best performance in previous experiments.
%To evaluate the generalization of the approach also to other views we tested using, DINOv2 and TAS model distillation, a second EXO view from the dataset (see Appendix~\ref{supp:view_selection} for more informations on view selection). In Table~\ref{tab:view} we show results for the second adaptation settings (v4$\to$e4). 
Exo oracle performance highlights that view $v4$ is less informative than $v3$ (compare line 1 of Table~\ref{tab:view} with line 2 of Table~\ref{tab:main_results} - e.g. Edit distance $24.03$ vs $28.55$).
Despite this, feature distillation allows to improve over the baseline (compare line 4 with line 3), increasing for instance edit score from $12.82$ to $20.24$ and MoF from $10.65$ to $24.18$. The TAS distillation module further improves edit score to $24.23$ (line 5), a result closer to the ego oracle (line 2, edit score of $26.42$). 
Enabling both feature and TAS distillation brings additional benefits in particular in the F1 and MoF metrics (compare line 6 with lines 4-5 and line 3), with overall improvements over the baselines of $+18.74$ in MoF, $+14.39$ in $F1_{25}$ and $+11.64$ in edit score.
%Despite this, our feature distillation followed by TAS model distillation allows to outperform the baseline (compare lines 3-6 of Table~\ref{tab:view}), with improvements of e.g., $+11.64$ in edit score, and reaches or even outperforms EXO oracle performance (compare lines 1-6 of Table - e.g., edit score of $24.03$ vs $24.46$ and MoF of $22.54$ vs $29.39$).
The findings indicate that when the EXO view provides less information, it's preferable to utilize feature distillation combined with TAS model distillation for DINOv2. 
%However, when the EXO view is more informative, TAS model distillation alone is sufficient to bridge the gap. 
%Overall, results show the effectiveness of the proposed model transfer approach both with domain-specific (TSM) and domain-agnostic (DINOv2) features, with domain-agnostic features being particularly promising for this task.

\subsection{Comparisons with the State of the Art} 
\label{sec:comparison_with_sota}
In Table~\ref{tab:synchronization}, we compare the proposed method with different state-of-the-art approaches for unsupervised domain adaptation and video alignment.
Note that, since no previous investigation has considered the proposed task of exo-to-ego transfer with unlabeled synchronized video pairs, we consider approaches working in the absence of synchronization. 
These comparisons also serve as a means to assess the contribution of synchronized video pairs to the adaptation process.
All these experiments are related to the $v3 \to e4$ adaptation scenario and use DINOv2 as feature extractor (our best performing configuration), performing TAS model distillation with different loss functions.
As domain adaptation approaches, we consider Maximum Mean Discrepancy (MMD)~\cite{gretton2012kernel}, Deep CORAL (DCORAL)~\cite{sun2016deep}, and Gradient Reversal Layer (GRL)~\cite{ganin2015unsupervised}. As video alignment losses, we consider Temporal Cycle Consistency (TCC)~\cite{dwibedi2019temporal} and Smooth Dynamic Time Warping (SDTW)~\cite{hadji2021representation}. We compare these approaches with a ``No Adaptation'' approach, which consists in testing the exocentric model directly on egocentric data, and with a ``Rnd. MSE'' baseline, which computes a random assignment between video frames and applies the MSE loss to random pairs. 
When training with the competitor losses, we feed two random exo and ego videos to assess the ability of the model to align features in an unsupervised way. In the case of alignment losses, we also assess the effect of feeding two videos of the same procedure (i.e., an assembly or disassembly of the same object) to ensure that an alignment between the videos indeed exists.

\begin{table}[t]
   
\centering
    %\captionsetup{font=scriptsize}  
    \caption{Comparison of the proposed synchronized model adaptation (last line) with different adaptation approaches. Methods denoted with * are trained on pairs of videos of the same procedure. Best results are in bold, second best results are underlined.}
   \vspace{2mm}
%\scriptsize
    \resizebox{0.6\linewidth}{!}{
    \begin{tabular}{crrrrr}
    %\hline
    \textbf{Adaptation} & \textbf{Edit} & $\bf{F1_{10}}$ & $\bf{F1_{25}}$ & $\bf{F1_{50}}$ & \textbf{MoF} \\
    \hline
    No Adaptation & 12.60 & 10.18 & 7.21 & 2.40 & 14.15 \\
    \hline
    MMD~\cite{gretton2012kernel} & 16.98 &	14.21 &	10.58 &	\underline{5.86} &	10.11 \\
    DCORAL~\cite{sun2016deep}  & 17.13 & \underline{16.25} & \underline{12.04} & 5.51 & \underline{17.72} \\
    GRL~\cite{ganin2015unsupervised} & 16.57 & 15.43 & 8.23 & 2.78 & 16.84 \\
    \hline
    TCC~\cite{dwibedi2019temporal} & 16.03 & 15.11 & 10.78 & 5.04 & 16.42 \\
    SDTW~\cite{hadji2021representation}  &  13.82 & 11.48 & 8.15 & 4.77 & 13.21 \\
    \hline
    TCC~\cite{dwibedi2019temporal}* &  \underline{17.84} &	15.91 &	11.59 &	5.48 &	17.63\\
    SDTW~\cite{hadji2021representation}*  & 15.27 &	12.22 &	9.19 &	4.94 &	14.82   \\
    \hline
    Rnd. MSE  & 13.27 & 10.03 & 6.98 & 2.36 & 10.61\\
    \hline
    Sync. (ours) &\textbf{{28.59}} & \textbf{{29.58}} &\textbf{{24.84}} &	\textbf{{16.38}} &	\textbf{{31.36}}  \\
    \hline
    Imp. vs $2^{nd}$ best &  +10.75 & +13.33 & +12.8 & +10.52 & +13.64\\
    \hline
    
\hline
\end{tabular}}
\label{tab:synchronization}

\end{table}

Performing distillation through a random assignment (Rnd. MSE) does not bring systematic improvements on No Adaptation (compare first line with third-last). Domain adaptation approaches (lines 2-4) bring improvements, with DCORAL obtaining an edit score of $17.13$ vs $12.60$ ($+4.53$) of No Adaptation. Improvements are less pronounced in the case of unsupervised alignment losses on random video pairs (lines 5-6), with TCC obtaining an edit score of $16.03$, a $+3.43$ over No Adaptation.
Considering video pairs of the same procedure significantly improves results of unsupervised alignment losses (lines 6-7), bringing TCC to the second best result of $17.84$ according to the edit score.
The proposed distillation approach (second-last line) improves over the second best results by $+10.75$ in terms of edit score and $+13.64$ in terms of $MoF$ (see last line). These results highlight the effectiveness of considering unlabeled synchronized video pairs as a source of supervision.

\begin{table}[t]
   
    \begin{minipage}[t]{0.49\textwidth}
    \begin{flushleft} 
    \centering
    %\captionsetup{font=scriptsize}
    \caption{Performance with different amounts of unlabeled synchronized Exo-Ego video pairs. The proposed method doubles results over the baseline with $50\%$ of the data, about $5$ hours of video.}
    \vspace{2 mm}
    \resizebox{\linewidth}{!}{
    \begin{tabular}{cc|rrrrr}
    %\hline
    \textbf{\% Pairs} & \textbf{\# Hours} & \textbf{Edit} & $\bf{F1_{10}}$ & $\bf{F1_{25}}$ & $\bf{F1_{50}}$ & \textbf{MoF} \\
    \hline
0\% & 0 & 12.60 & 10.18 & 7.21 & 2.40 & 14.15\\ 
\hline
%5\% & 0.56 & 13.19 & 8.83 & 6.79 & 4.04 & 10.22\\    
10\% & 1.04 & 13.58 & 9.21 & 7.78 & 4.15 & 10.75\\    
15\% & 1.43 & 16.31 & 15.52 & 12.4 & 7.15 & 17.27\\    
20\% & 2.09 & 19.85 & 18.34 & 15.31 & 10.95 & 21.32\\    
25\% &	2.78 &	22.73 &	23.52 &	19.29 &	12.04 &	26.11 \\
50\% &	5.23 &	25.06 &	25.62 &	21.07 &	13.06 &	29.00 \\
75\% &	8.06&	25.14&	26.41&	21.87	&15.39&	29.55 \\
\hline
100\% &	10.21&	28.59&	29.58&	24.84&	16.38 &	31.36 \\
\hline
\end{tabular}}
\label{tab:amounts_synchronized}
\end{flushleft}
\end{minipage}
\hfill
\begin{minipage}[t]{.49\linewidth}
\begin{flushright}
\centering
    \caption{\label{Jitter} 
DINOv2 model performance with different drop rates. The method shows good performace for small drops, $0.5\%$, and double baseline performance with a drop of $2.0\%$}
\vspace{2 mm}
    \resizebox{\linewidth}{!}{
\begin{tabular}{c|ccccc}

\textbf{Drop Rate} & \textbf{Edit} & $\bf{F1_{10}}$ & $\bf{F1_{25}}$ & $\bf{F1_{50}}$ &  \textbf{MoF} \\ \hline
 
%6 shift & 27.31 & 27.85 & 22.85 & 14.53 & 30.99\\ 
%Black &  &  &  &  & \\ \hline
No Drop & 28.59 & 29.58 & 24.84 & 16.38 & 31.36 \\
\hline
$0.5\%$ & 27.68 & 27.87 & 23.84 & 16.29 & 31.22 \\
$1.5\%$ & 26.91 & 27.56 & 23.86 & 16.23 & 30.83\\ 
$2.0\%$ & 26.17 & 26.38 & 22.23 & 14.74 & 29.69\\ 
$2.5\%$ & 24.63 & 24.05 & 20.25 & 14.13 & 28.16\\ 
$3.0\%$ & 22.19 & 21.15 & 17.98 & 10.92 & 25.63\\ 
%$3.5\%$ & 17.36 & 15.91 & 11.39 & 5.13 & 17.03\\ 
\hline
No adaptation & 12.60 & 10.18 & 7.21 & 2.40 & 14.15 \\
\hline

\end{tabular}
}
\end{flushright}
\end{minipage}
\end{table}

\subsection{Amounts of Video Pairs and Accuracy of Synchronization} Table~\ref{tab:amounts_synchronized} reports the performance of our adaptation method based on DINOv2 when different amounts of unlabeled synchronized Exo-Ego video pairs are used for knowledge distillation. Our model doubles results over the baseline with about $5$ hours of video. This highlights the convenience of the proposed approach even in those scenarios in which it is possible to collect only a limited number of synchronized Exo-Ego video pairs.

We further assess the robustness of our method in the presence of imperfect synchronization, which can arise from faulty sensors or unsophisticated synchronization approaches.
We trained our DINOv2-based model on $v3 \to e4$ randomly dropping a given percentage of frames independently from the paired Ego and Exo videos, thus affecting their synchronization.
Table~\ref{Jitter} reports results for different drop rates, demonstrating that our method is resilient to small drop rates, despite not being explicitly designed for this. For reference, a drop rate of $3\%$ implies that $\sim1$ frame/sec is dropped at $30fps$. Even with this significant amount of noise, the proposed adaptation offers advantages over the ``no adaptation'' baseline.
See the supplementary material for additional experiments on training data and feature extractors.

\subsection{Analysis on the EgoExo4D Dataset}
Table~\ref{tab:egoexo4d} reports results on EgoExo4D. Oracle performance shows that, in this setting, the EGO signal (line 2 Edit $28.74$) is stronger than the EXO signal (line 1 Edit $21.64$).
We believe that this depends on the high variability in terms of environments and view points of the exocentric cameras which are not stationary across videos. 
As a result, the gap between a no-adaptation baseline (line 3) and the oracle (line 1) is limited ($20.56$ vs $21.64$). 
Despite this small gap, the distillation process (line 4) improves results over the baseline (line 3) with improvements of $+3.32$ in Edit, $+5.91$ in MoF, and other positive increments in the other measures.
Remarkably, the distillation process even surpasses the oracle ($23.88$ edit in line 4 vs $21.64$ in line 1), which is due to the higher quality of information captured by the egocentric view.
Given the advantage of the egocentric view in this context, we also experiment with an $EXO \to EGO$ adaptation, in which an EGO teacher is distilled into a model aiming to improve performance in the exocentric domain.
Our methodology generalizes also to this case, with the adapted TAS model (line 7) achieving an edit score of $20.49$, an improvement of $+4.44$ over the no adaptation baseline (line 6), reaching a result comparable to the EXO oracle (line 1 - $21.64$ edit).

%, a single model that efficiently classifies images, videos, and single-view 3D data with the same set of model parameters 
%It takes advantage of transformer-based architectures' flexibility and is trained simultaneously on classification tasks covering various modalities. The common visual representation employed by Omnivore simplifies cross-modal recognition without requiring correspondences between different modalities.
%using the following setup: features are derived for every take, camera, and camera stream, a stride of $16/30$ seconds is employed, accompanied by a window size of $32/30$ seconds.
%In cases where the stride does not evenly divide the total duration time, the last [$n - 32/30$, n) seconds of the video are utilized as the final window.

\begin{table}[t]
    \begin{minipage}[b]{0.49\linewidth}
        \centering
        \caption{Results on EgoExo4D~\cite{grauman2023ego}. \faEyeDropper denotes TAS model distillation. Best results within the same group are highlighted.}
        \vspace{2mm}
    \resizebox{\linewidth}{!}{
    \begin{tabular}{r|c|c|rrrrr}
    \multicolumn{1}{c}{} & \multicolumn{1}{c}{\textbf{Adaptation Task}}  &   \multicolumn{1}{c}{\textbf{\faEyeDropper}} & \multicolumn{1}{c}{\textbf{Edit}} & \multicolumn{1}{c}{$\bf{F1_{10}}$} & \multicolumn{1}{c}{$\bf{F1_{25}}$} & \multicolumn{1}{c}{$\bf{F1_{50}}$} & \multicolumn{1}{c}{\textbf{MoF}} \\
    \cline{1-8}
    1&EXO $\to$ EXO  & & 21.64 &	16.40	&11.69	&05.89	&24.17 \\
    \cline{1-8}
    2&EGO $\to$ EGO  &&28.74	&24.55	&19.88	&10.21	&40.45   \\
    \hline
    3&\multirow{2}{*}{EXO $\to$ EGO}  & & 20.56 & 16.15 & 11.20 & 05.24 & 24.81 \\
    4& &   \checkmark & \textbf{23.88} & \textbf{18.12}  &  \textbf{12.20} &	\textbf{05.50} & \textbf{30.72}  \\
    
    \cline{3-8}
    \hline
    5&\multicolumn{2}{c|}{impr. w.r.t. line 3}&  {$+3.32$}& {$+1.97$}& {$+1.00$}& {$+0.26$}&{$+5.91$}   \\
    \hline
    6&\multirow{2}{*}{EGO $\to$ EXO}  & &  16.05&	14.64&	10.60&	04.82&	\textbf{26.82}  \\
    7&&   \checkmark &  \textbf{20.49}& \textbf{15.62}& \textbf{10.98}& \textbf{04.90}&25.67   \\
    \hline
    8&\multicolumn{2}{c|}{impr. w.r.t. line 6}& {$+4.44$}& {$+0.98$}& {$+0.38$}& {$+0.08$}&{$-1.15$}   \\
    \hline
    \cline{1-8}
    
\end{tabular}}
\label{tab:egoexo4d}  
\vspace{4mm}
    \end{minipage}\hfill%
    \begin{minipage}[b]{0.49\linewidth}
        \centering
        \includegraphics[width=\linewidth]{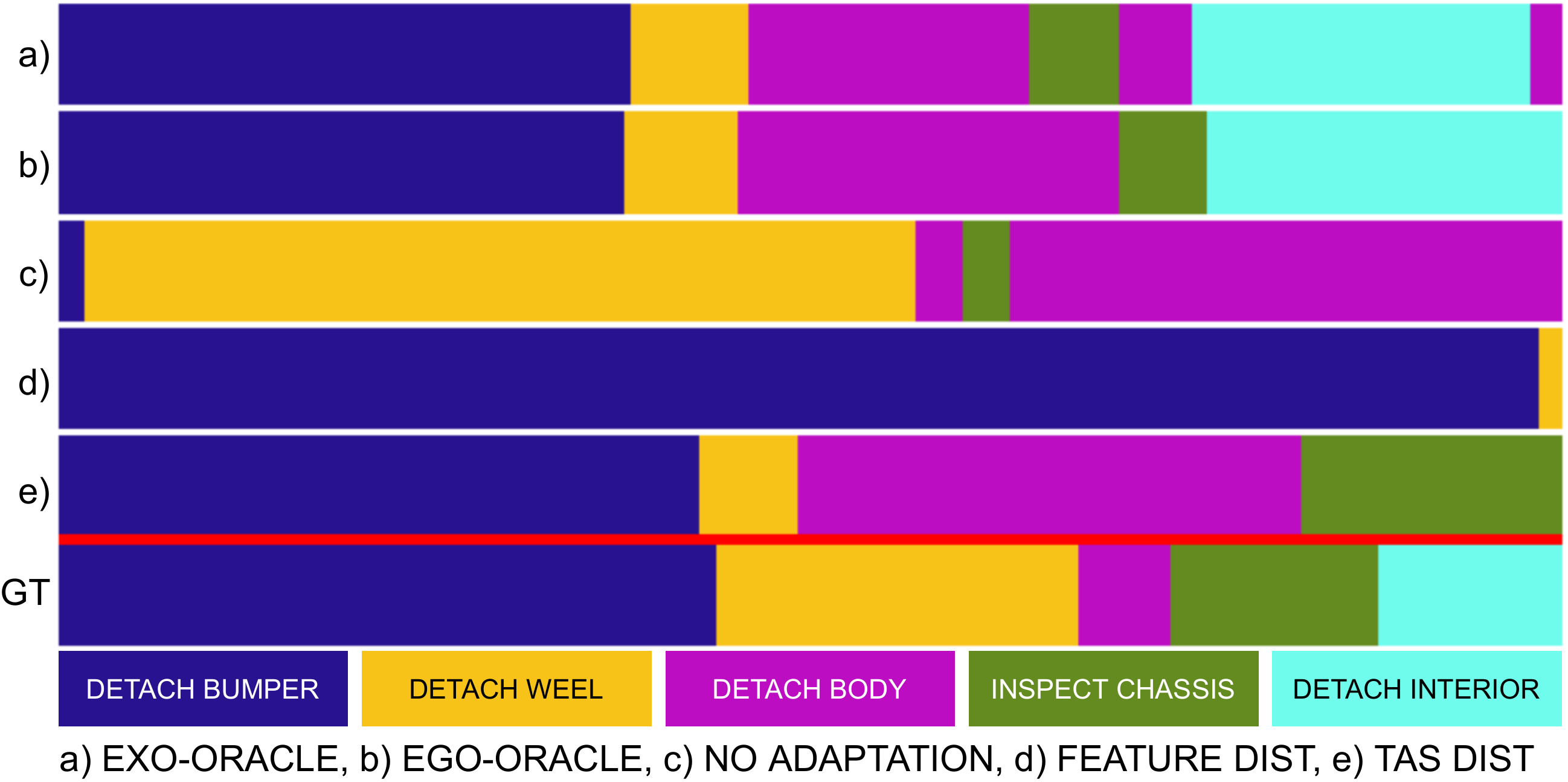}
        \captionof{figure}{Qualitative example of temporal segmentations related to different approaches on a test video.}
        \label{fig:qualitative}

    \end{minipage}%
    %\vspace{-1em}
\end{table}

\subsection{Qualitative Results}
\Cref{fig:qualitative} reports qualitative segmentations of the compared approaches on a test video. Compared methods are based on the DINOv2 feature extractor.
%The most accurate predictions are obtained by EXO-ORACLE and EGO-ORACLE, which retrieve most of the segments and predict a likely action sequence. 
%The NO-ADAPTATION baseline achieves poor performance, under-segmenting the video and predicting inaccurate temporal bounds.
%TAS MODEL DIST. retrieves most ground truth segments and returns a likely sequence of actions, missing only the last ``detach interior'' step. 
More examples and discussion in supplementary material.
%\footnote{More examples are reported in Appendix ~\ref{supp:qualitative_examples} of the supp.}

%\begin{figure}[t!]
%    \centering
    %\vstretch{.7}
%    {\includegraphics[width=0.69\linewidth]{imgs/qualitative.pdf}}
%    \caption{Qualitative example of temporal segmentations related to different approaches on a test video.}
%    \label{fig:qualitative}
%\end{figure}

\section{Conclusion}
We have shown that temporal action segmentation models can be transferred from the exocentric to the egocentric domain, provided that labeled exocentric data and unlabeled synchronized exocentric-egocentric video pairs are given.
%We proposed an adaptation model based on knowledge distillation which, 
Without bells and whistles, our approach performs on par with fully supervised approaches trained on labeled egocentric data,  scoring a $+15.99$ improvement in the edit score on Assembly101  and $+3.32$ on the challenging EgoExo4D dataset compared to a baseline trained on exocentric data only. 
%The superiority of the proposed approach has been evaluated with respect to several methods from the unsupervised domain adaptation and temporal sequence alignment literature. 
Code is available here: \url{https://github.com/fpv-iplab/synchronization-is-all-you-need}.

\section*{Acknowledgements}
This research has been supported by the project Future Artificial Intelligence Research (FAIR) – PNRR MUR Cod. PE0000013 - CUP: E63C22001940006

%\par\vfill\par
%Now we have reached the maximum length of an ECCV \ECCVyear{} submission (excluding references).
%References should start immediately after the main text, but can continue past p.\ 14 if needed.
%\clearpage  % TODO REVIEW/FINAL: This \clearpage needs to be removed from both review and camera-ready versions.


\bibliographystyle{splncs04}
\bibliography{main}

%\section{Supplementary Material}
%\subsection{Details on Feature Distillation}
%\label{supp:backbone_distillation}

\clearpage
\appendix
\setcounter{page}{1}
\title{Synchronization is All You Need: Exocentric-to-Egocentric Transfer for Temporal Action Segmentation with Unlabeled Synchronized Video Pairs\\(Supplementary Material)}

\author{Camillo Quattrocchi\inst{1}\orcidlink{0000-0002-4999-8698} \and
Antonino Furnari\inst{1,2}\orcidlink{0000-0001-6911-0302} \and
Daniele Di Mauro\inst{2}\orcidlink{0000-0002-4286-2050}  \and \\
Mario Valerio Giuffrida\inst{3}\orcidlink{0000-0002-5232-677X} \and
Giovanni Maria Farinella\inst{1,2}\orcidlink{0000-0002-6034-0432}}

\authorrunning{C.~Quattrocchi et al.}
\titlerunning{Synchronization is All You Need}

\institute{Department of Mathematics and Computer Science, University of Catania, Italy \and
Next Vision s.r.l., Italy \and
School of Computer Science, University of Nottingham, United Kingdom}

\maketitle 
\section{Details on Feature Distillation}
\label{supp:backbone_distillation}

\begin{figure*}[!t]
    \centering
    \begin{subfigure}{.45\textwidth}
        \includegraphics[width=\textwidth]{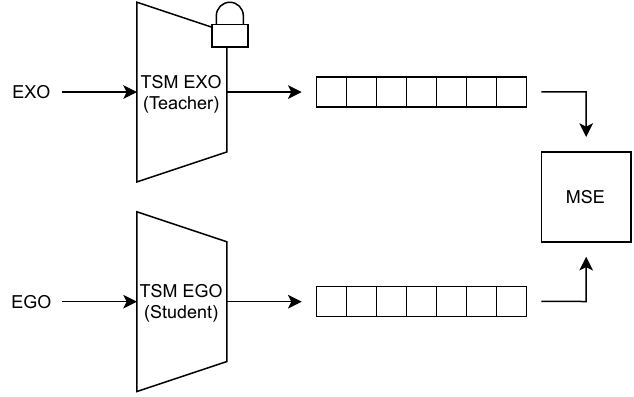}
        \caption{Adaptation of TSM feature extractor.}
        \label{supp:feature_distillation}
    \end{subfigure}\quad
    \begin{subfigure}{.45\textwidth}
        \includegraphics[width=\textwidth]{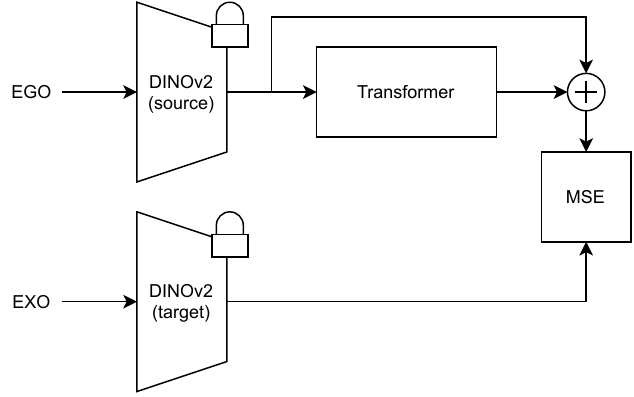}
        \caption{Adaptation of DINOv2 feature extractor.}
        \label{supp:feature_distillation_dino}
    \end{subfigure}\quad
\caption{Features Distillation processes.}
    \label{img:distillation_processes}
\end{figure*}

\Cref{img:distillation_processes} shows a scheme of the feature distillation process. The process aims to bring exocentric representations as close as possible to egocentric ones. Specifically, \Cref{supp:feature_distillation} shows the distillation process based on the TSM feature extractor. \textcolor{red}{} The two TSM models are pretrained on EPIC-KITCHENS-100. The teacher model is trained using third-person videos. During the distillation phase, the two models see synchronized Exo-Ego clips. The teacher model is frozen and the student model is trained to predict features similar to the ones of the Exo model through an MSE loss as denoted in the following:

\begin{equation}
\mathcal{L}_f=\frac{1}{M}\sum_{j=1}^{M}||\phi^S(S_j^{ego}) - \phi^T(S_j^{exo})||^2    
\end{equation}

\noindent
where $\mathcal{L}_f$ is the feature distillation loss, $M$ is the number of examples in a batch, $\phi^S$ and $\phi^M$ are the student and teacher TSM models respectively, and $S_j^{ego}$ and $S_j^{exo}$ are the synchronized egocentric and exocentric video pairs.

\Cref{supp:feature_distillation_dino} shows the distillation process based on DINOv2 features. We implement $\theta$ as a residual transformer encoder network that translates video feature sequences in one pass. The transformer network has $12$ layers, $16$ heads, and a number of hidden units equal to $1024$. At test time, the adapted DINOv2 features are extracted by the composed function $\theta(\phi(\cdot))$. The adapter module is trained to make DINOv2 features extracted from Ego videos closer to the DINOv2 features extracted from Exo features. The transformer module is fine-tuned during model distillation. Note that we do not fine-tune DINOv2 model due to the large amount of computational resources that such a fine-tuning would need. Being DINOv2 trained on huge quantities of data, we would also expect it to easily overfit on Assembly101. Also in this case, the model is trained using an MSE loss.

\begin{figure}
    \centering
    \includegraphics[width=\linewidth]{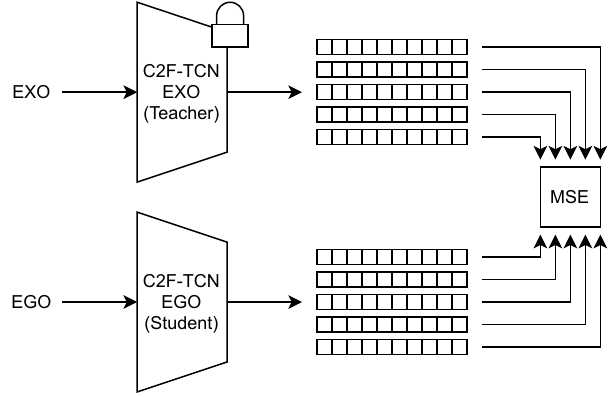}
    \caption{TAS Model distillation process.}
    \label{supp:model_distillation_im}
\end{figure}

\begin{figure*}[!t]
  \centering
   \includegraphics[width=\linewidth]{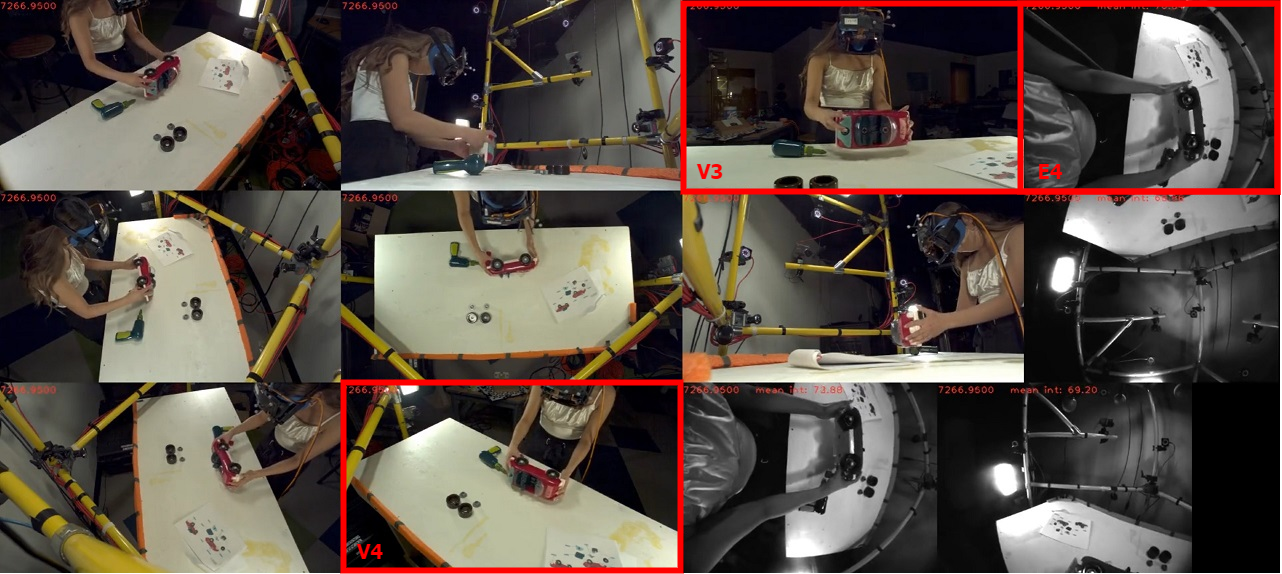} 
   \caption{Assembly101 views. The views chosen for our experiments are highlighted in red.}
   \label{fig:selected_views}
\end{figure*}

\section{Details on TAS Model Distillation}
\label{supp:model_distillation}
Figure~\ref{supp:model_distillation_im} shows a scheme of the TAS model distillation process. In this scenario, the teacher model is obtained by training a C2F-TCN model on a collection of features extracted from labeled exocentric videos ($\mathcal{D}_{train}^{exo}$). The student network is a different instance of C2F-TCN. During the TAS model distillation phase, the two models take as input synchronized exocentric-egocentric video pairs ($\mathcal{D}_{pair}^{adapt}$). Each of the two C2F-TCN models produce hierarchical representations with different temporal resolutions. We use an MSE loss to align features at the different levels. Note that, as in previous distillation schemes, also in this case the teacher is fixed. The total loss is obtained as the sum of each individual MSE loss function calculated for each temporal resolution, as denoted in the following:

\begin{equation}
\mathcal{L}_m = \frac{1}{M}\sum_{j=1}^M \sum_{l=1}^L ||\gamma^S_l(\phi^S(V_j^{ego})) - \gamma^T_l(\phi^T(V_j^{exo}))||^2
\end{equation}

\noindent
where $\mathcal{L}_m$ is the TAS model distillation loss, $\phi(V)$ is a sequence of frame-wise features extracted from $V$, $M$ is the number of examples in a batch, $L$ is the number of layers in the C2F-TCN model, $\gamma_l^S$ and $\gamma_l^T$ are the feature extracted at the $l^th$ level of the feature hierarchy by the student and teacher models respectively, $V_j^{ego}$ and $V_j^{exo}$ are the paired Ego and Exo videos. 

Our TAS model distillation loss aligns features at all levels of the hierarchy. We validated this design choice by applying the loss function to different sets of layers. Results reported in~\Cref{tab:different_layers} show that aligning features at all levels brings best performance.

\begin{table}[t]
\caption{Results obtained applying TAS model distillation at different sets of layers.}
\centering
\resizebox{0.7\linewidth}{!}{
\begin{tabular}{c|lllll}
\multicolumn{1}{l|}{Layers} & \textbf{Edit}  & $\bf{F1_{10}}$ & $\bf{F1_{25}}$ & $\bf{F1_{50}}$ & \textbf{MoF}   \\ \hline
1                           & 23.45          & 23.36             & 19.60              & 12.84             & 28.83          \\
1-2                           & 26.45          & 28.69             & 23.67             & 15.42             & 31.66          \\
1-3                           & 26.97          & 26.67             & 23.10              & 15.43             & 31.20           \\
1-4                           & 27.02          & 27.33             & 23.09             & 15.89             & 31.17          \\
\textbf{1-5}                  & \textbf{28.59} & \textbf{29.58}    & \textbf{24.84}    & \textbf{16.38}    & \textbf{31.36}\\
\hline
\end{tabular}}
\label{tab:different_layers}
\end{table}

\section{Details on Assembly101 - View Selection}
\label{supp:view_selection}

As already discussed in the main paper, the Assembly101 dataset was acquired using 4 first-person monochrome cameras mounted on a custom headset, with a resolution of $640\times480$ pixels, and 8 third-person RGB cameras (5 mounted overhead, 3 on the side), with a resolution of $1920\times 1080$ pixels. \Cref{fig:selected_views} shows all the views for an example frame of the dataset with the selected ones marked in red. 
We select a representative egocentric view and two exocentric views to obtain the two video pairs considered in the paper.
Two of the egocentric views point towards the ceiling, while the remaining two are redundant. We choose one of the two informative one, $e4$, as a representative egocentric view.
We the select $v4$ and $v4$ as two representative exocentric views showing the scene from typical, but different points of views.
%We selected the exocentric view to minimize the number of occlusions as the operator performs a given task.
%This view is representative of a third-person vision system in which the camera is placed in order to best acquire fine-grained details on the executed procedures.
%The egocentric view is chosen to maximize the amount of procedure-related captured visual content. Note that, among the $4$ available egocentric views, two are upward-looking, capturing only static elements of the scene such as the ceiling, so they have been discarded. The remaining two views represent two typical first-person views and are very overlapped, hence capturing similar visual information. We chose one of the two as a representative.

\section{Training Details}
\label{supp:training_details}
All the experiments have been performed with the following setup. Teacher and student TSM models are pre-trained on EPIC-KITCHENS-100~\cite{Damen2020RESCALING} as specified in~\cite{sener2022assembly101}. We use the code provided by the authors of~\cite{sener2022assembly101} to train the TSM and C2F-TCN models. We use the \textit{large} DINOv2 model to extract per-frame features using the official implementation\footnote{\url{https://github.com/facebookresearch/dinov2}}. The Learning rate and weight decay were set at $10^{-4}$ and we adopted ADAM optimizer for training. To avoid overfitting, we adopted an early stop criterion, with the maximum number of epochs equal to 200 and a patience of 20. This criterion involves halting the training process before completion based on the model's performance on the training data. This practice prevents overfitting by stopping training when the performance ceases to improve. The transformer learning rate was initialized to $10^{-2}$ and we adopted SGD optimizer for training. During the experiments presented in the main paper, the classification version of the TCC loss was used. 

\section{Ablation on the use of Exo and Ego data to train the feature extractor on Assembly101}
\label{supp:ablate_exo_ego_data}
\begin{table}[t]
   
    \centering
    %\captionsetup{font=scriptsize}
    \caption{Effect of using exo (\faCamera) and ego (\faGlasses) data to train the TSM backbone. Using both exo and ego data brings little benefits when testing the segmentation model on exocentric data (rows 1-2). Clearer benefits are obtained when models are tested on egocentric data (lines 3-6). Feature + TAS model distillation (\faEyeDropper), in the last line, has a similar effect to using ego labeled data for training (last line vs line 3). Best results within the same group are highlighted in bold.}
    \vspace{0.1 mm}
\resizebox{0.7\linewidth}{!}{
    \begin{tabular}{ccccrrrrr}
    %\hline
    \textbf{Task} & \textbf{\faCamera} & \textbf{\faGlasses} & \textbf{\faEyeDropper} & \textbf{Edit} & $\bf{F1_{10}}$ & $\bf{F1_{25}}$ & $\bf{F1_{50}}$  & \textbf{MoF} \\
    \hline
    \multirow{2}{*}{\makecell{EXO $\to$ EXO\\(exo-oracle)}} & \checkmark & & &\textbf{31.45}                     & 34.24                     & \textbf{29.92}                     & \textbf{20.88}                     & 37.39                     \\
&\checkmark &\checkmark& &31.14& \textbf{34.95} & 29.39 & 20.18 & \textbf{38.23} \\
\hline
\multirow{2}{*}{\makecell{EGO $\to$ EGO\\(ego-oracle)}} &&\checkmark& &29.25                     & 31.06                     & 25.40                     & 17.47                     & 36.60                     \\
& \checkmark&\checkmark& &\textbf{30.50}                     & \textbf{32.53}                     & \textbf{28.18}                     & \textbf{19.70}                     & \textbf{38.00}                     \\
\hline
\multirow{3}{*}{EXO $\to$ EGO} &\checkmark & &&10.53                    & 6.47                     & 2.44                     &  0.56                    &    5.42                  \\

&\checkmark &\checkmark& &\textbf{28.28}                     & \textbf{30.13}                     & \textbf{24.96}                     & \textbf{16.81}                     & \textbf{35.37}                     \\
& \checkmark & & \checkmark&{25.39}	&{26.79}	&{22.37}	&{15.52}	&{30.30}    \\
\hline
    \end{tabular}
    }
    \label{tab:egoexo_results}

\end{table}
\Cref{tab:egoexo_results} reports the results of experiments aimed to assess the effect of using both exo and ego data to train the TSM feature extractor, in comparison with distillation.
We consider four scenarios: exo-oracle (lines 1-2), ego-oracle (lines 3-4), exo-to-ego transfer without adaptation (lines 5-6), and ego-to-exo transfer with feature and TAS model distillation (line 7).
Using both egocentric and exocentric supervised data to train the TSM feature extraction module does not bring significant benefits in exo-oracle (comparing rows 1-2, edit scores of $31.14$ vs $31.45$ and similar MoF scores of $32.23$ vs $37.39$).
This suggests that exo videos provide a stronger signal, hence there is no benefit in adding egocentric supervisory signals when the system has to work in an exo setting.
Some benefits are obtained when exo data is added to an ego-oracle baseline (lines 3-4), with ego-oracle improving when both exo and ego data are used to train the feature extractor (edit score of $30.50$ vs $29.25$ and MoF of $38.00$ vs $36.60$).
This suggests that exocentric videos provide stronger signals which can regularize learning even when the model has to be tested on egocentric videos.
Using ego data to train the feature extractor brings significant improvements in the adaptation scenario: comparing line 6 with line 5, we observe that adding ego training data brings the edit score from $10.53$ to $28.28$, the $F1_{50}$ from $0.56$ to $16.81$ and the MoF from $5.42$ to $35.37$. These results suggest that the main limitation of the baseline trained only on exo data (line 5) lies in the limited expressiveness of exocentric features. It is worth noting that adding exo data to train the feature extractor allows the model to achieve similar performance to ego-oracle (comparing lines 6 vs 3, $28.28$ vs $29.25$ edit distance and $35.37$ vs $36.6$ MoF). Performing TAS model distillation has a similar effect, narrowing down the exo-ego gap and bringing performance closer to the ego-oracle level (comparing line 7 vs 3, $25.39$ vs $29.25$ edit score and $15.52$ vs $17.47$ $F1_{50}$), which further confirms the effectiveness of the proposed distillation approach.

\section{Ablation on the effect of different feature extractors on Assembly101 }
\label{sup:ablate_feat_ext}
\begin{table}[t]
    \centering
    %\captionsetup{font=scriptsize}
    \caption{Effect of different feature extractors when the segmentation model is trained and tested on egocentric data. Domain agnostic DINOv2 features outperform EXO features (line 2 vs line 1). Performing feature distillation (\faEyeDropper) improves the results (line 3 vs lines 1-2) and achieves performance similar to the backbone trained on EGO features (line 3 vs line 4). Best results are reported in bold.}
    \vspace{6 mm}
    \resizebox{0.7\linewidth}{!}{
    \begin{tabular}{cccrrrrr}
    %\hline
    \textbf{Backbone} & \textbf{Train} & \textbf{\faEyeDropper} & \textbf{Edit} & $\bf{F1_{10}}$ & $\bf{F1_{25}}$ & $\bf{F1_{50}}$ & \textbf{MoF} \\
    \hline
    TSM                                   & EXO                                &                                                        & 22.83                     & 20.57                     & 17.05                     & 10.92                     & 24.56                     \\
DINOv2                                   &  /                           &                               &  26.42 &26.50&22.16&18.84&33.47\\
TSM           &                      EXO                                 & \checkmark                       & 28.61 & 30.60 & \textbf{26.39} & \textbf{18.87}& 36.06 \\

TSM             &             EGO         &                               & \textbf{29.25}                     & \textbf{31.06}                     & 25.40                     & 17.47                     & \textbf{36.60}                     \\
\hline
\end{tabular}}
\label{tab:feat_ext_results}

\end{table}
\Cref{tab:feat_ext_results} reports the results of experiments aimed at assessing the effect of different feature extractors. For fair comparisons, for all approaches, temporal action segmentation models have been trained and tested using egocentric data, regardless of the data used to train the corresponding feature extractor, leading to a set of egocentric oracles, which differ only on the feature extractors. 
Methods obtained by training a TSM model on supervised exocentric data only achieve the worst performance (e.g., line 1 obtains an edit score of $22.83$).
Models based on the domain-agnostic DINOv2 feature extractor obtain better results (edit score of $26.42$ vs $22.83$ and an MoF of $33.47$ vs $24.56$ when comparing line 2 to line 1). 
The best overall results are obtained by an ego-oracle training the TSM module directly on egocentric features (line 4 obtained an edit score of $29.25$). 
Interestingly, distilling the exocentric model with unlabeled synchronized exo-ego data allows to achieve similar performance (line 4 has an edit score of $28.61$ and even outperforms the ego-oracle - line 3 - w.r.t. $F1_{25}$ and $F1_{50}$ measures), which highlights the ability of the feature distillation process to partially fill the representation gap.

\section{Additional Qualitative Results on Assembly101}
\label{supp:qualitative_examples}

\Cref{img:qualitative_supp_DINO} and~\ref{img:qualitative_supp_TSM} report additional qualitative results. For each example, we report temporal action segmentation results for the compared methods, the ground truth segmentation, and a color-coded legend indicating the different action classes involved in the specific video.

\subsection{DINOv2}
\begin{figure*}[t]
    \centering
    \begin{subfigure}{.45\textwidth}
        \includegraphics[width=\textwidth]{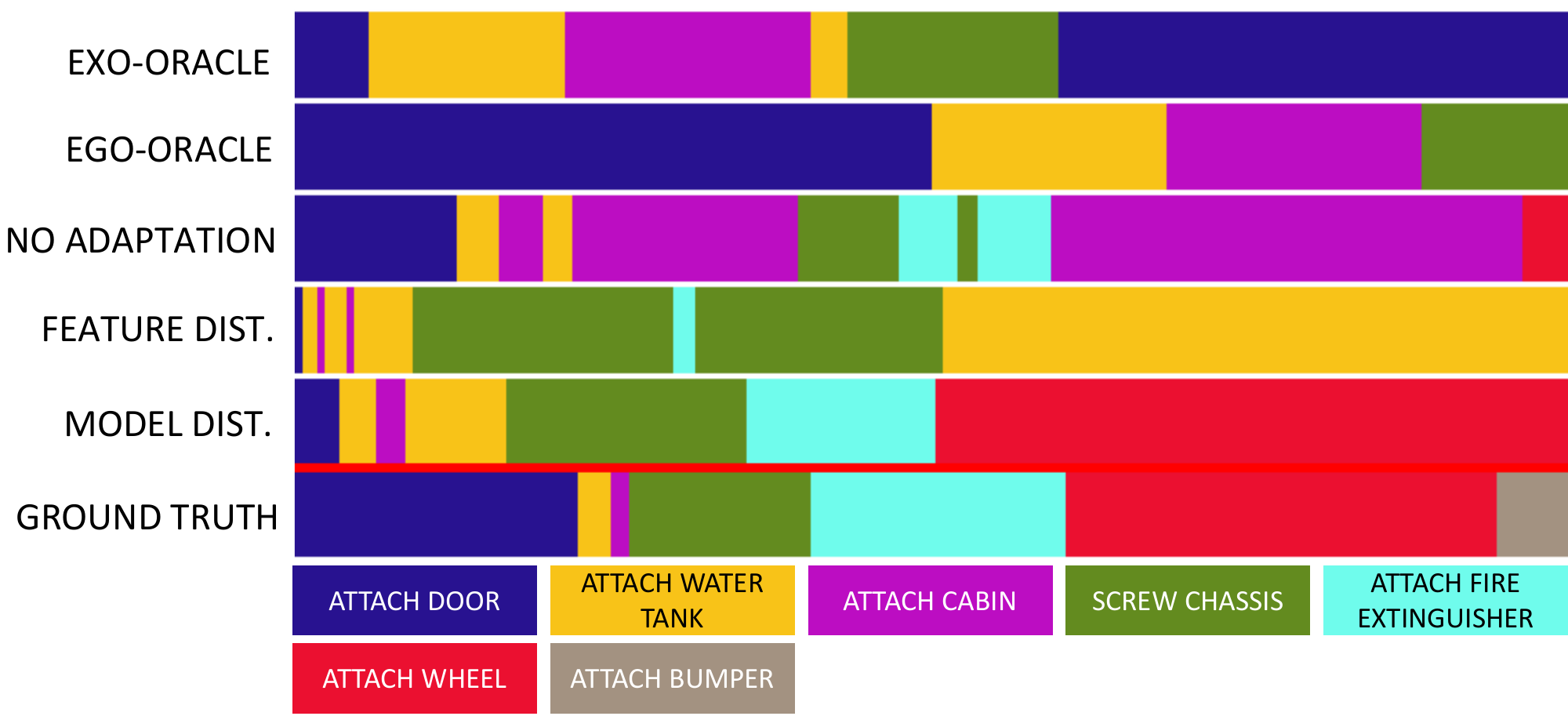}
        \caption{}
        \label{sub:dino_1}
    \end{subfigure}\quad
    \begin{subfigure}{.45\textwidth}
        \includegraphics[width=\textwidth]{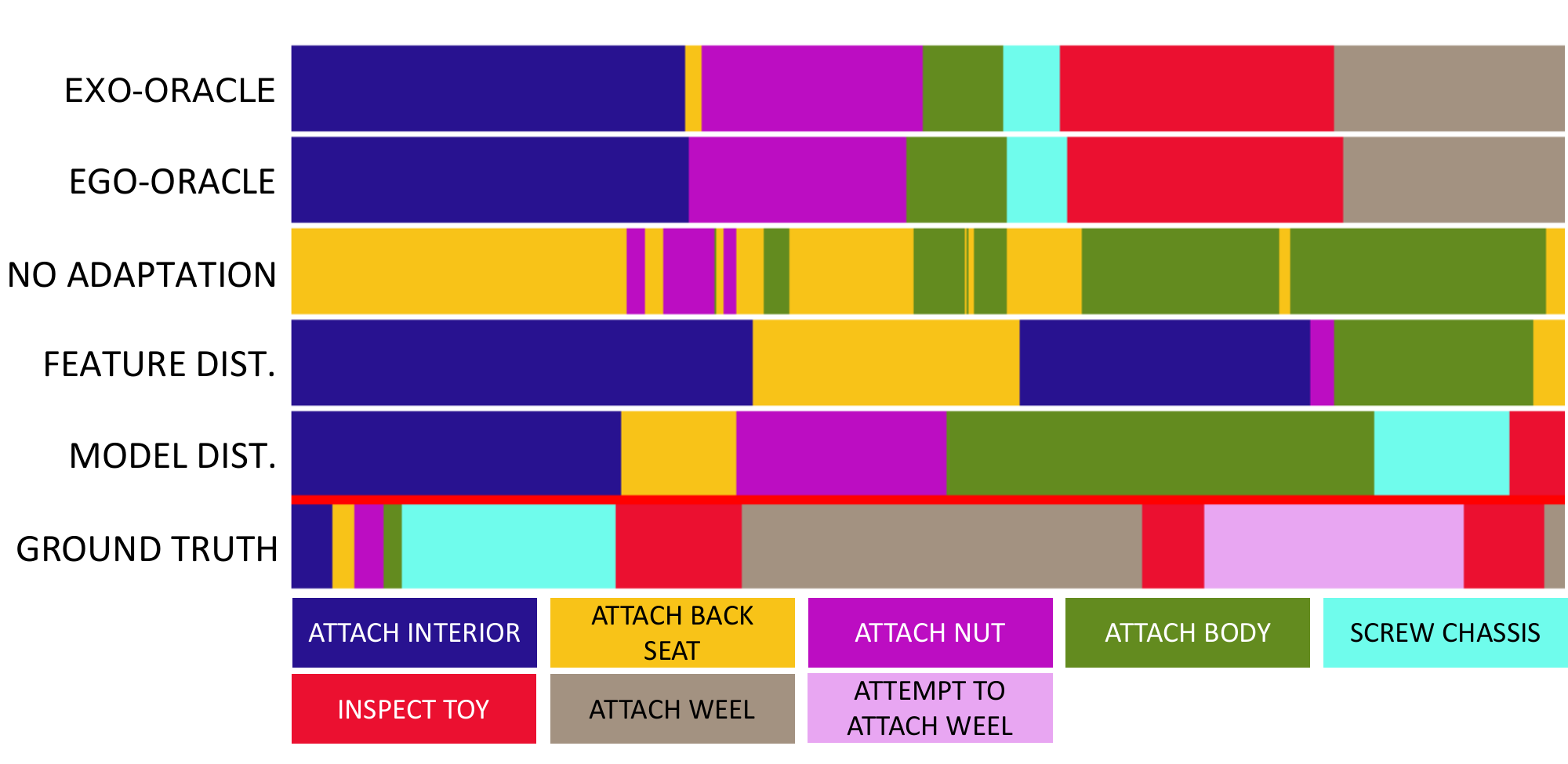}
        \caption{}
        \label{sub:dino_2}
    \end{subfigure}\quad
    \medskip
    \begin{subfigure}{.45\textwidth}
        \includegraphics[width=\textwidth]{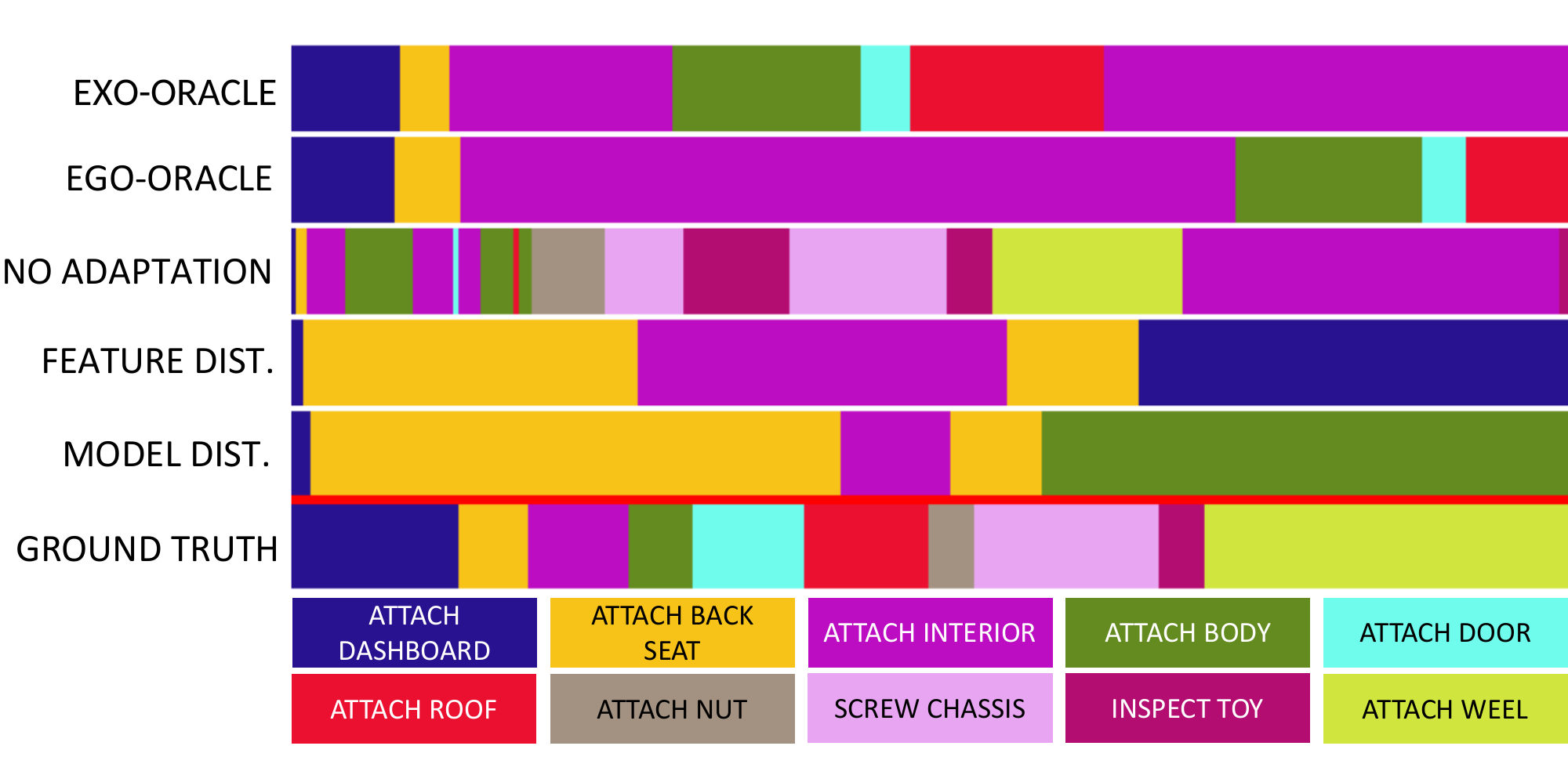}
        \caption{}
        \label{sub:dino_3}
    \end{subfigure}\quad
    \begin{subfigure}{.45\textwidth}
        \includegraphics[width=\textwidth]{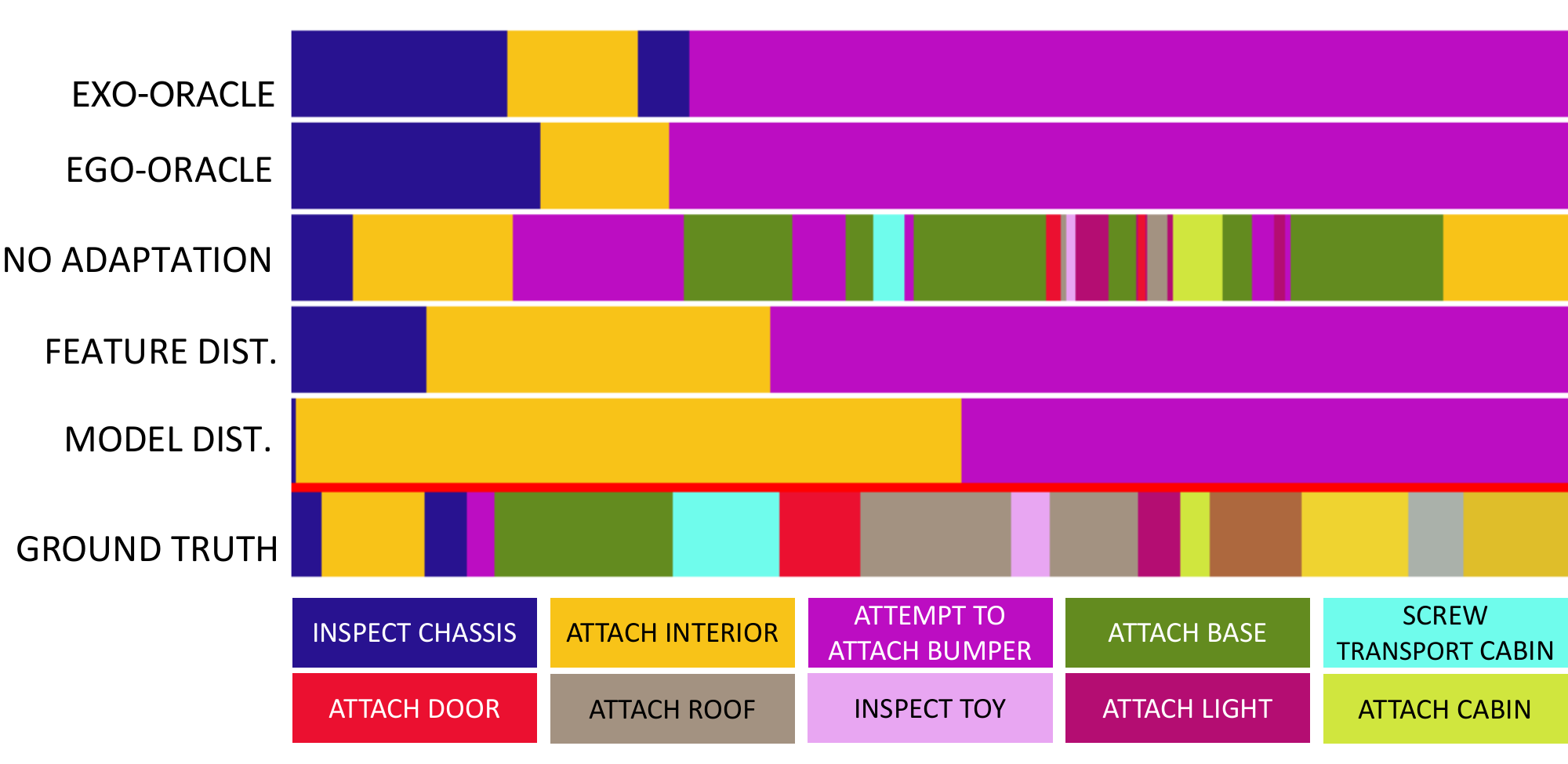}
        \caption{}
        \label{sub:dino_4}
    \end{subfigure}\quad
    \medskip
    \begin{subfigure}{.45\textwidth}
        \includegraphics[width=\textwidth]{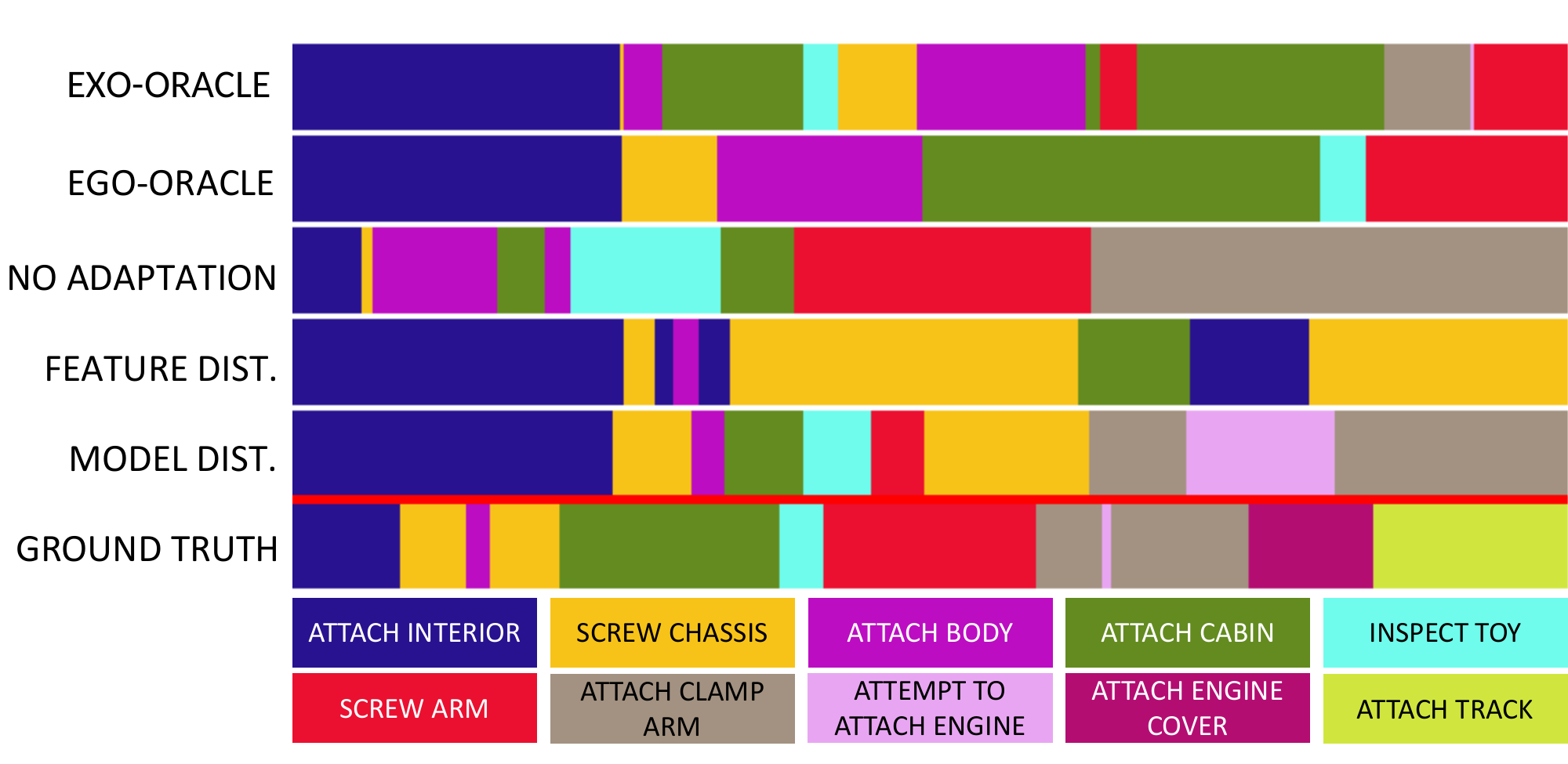}
        \caption{}
        \label{sub:dino_5}
    \end{subfigure}\quad
    \begin{subfigure}{.45\textwidth}
        \includegraphics[width=\textwidth]{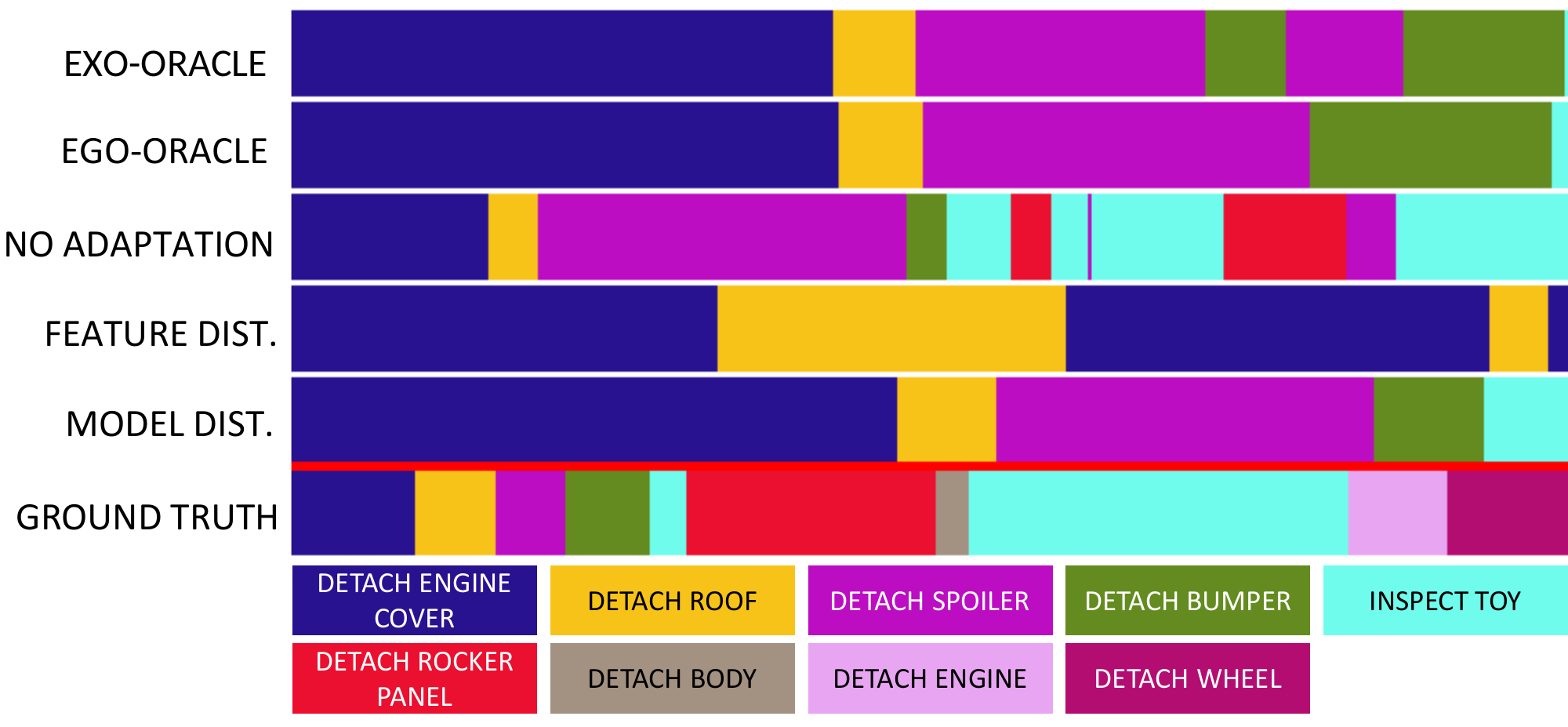}
        \caption{}
        \label{sub:dino_6}
    \end{subfigure}\quad
    \medskip
    \begin{subfigure}{.45\textwidth}
        \includegraphics[width=\textwidth]{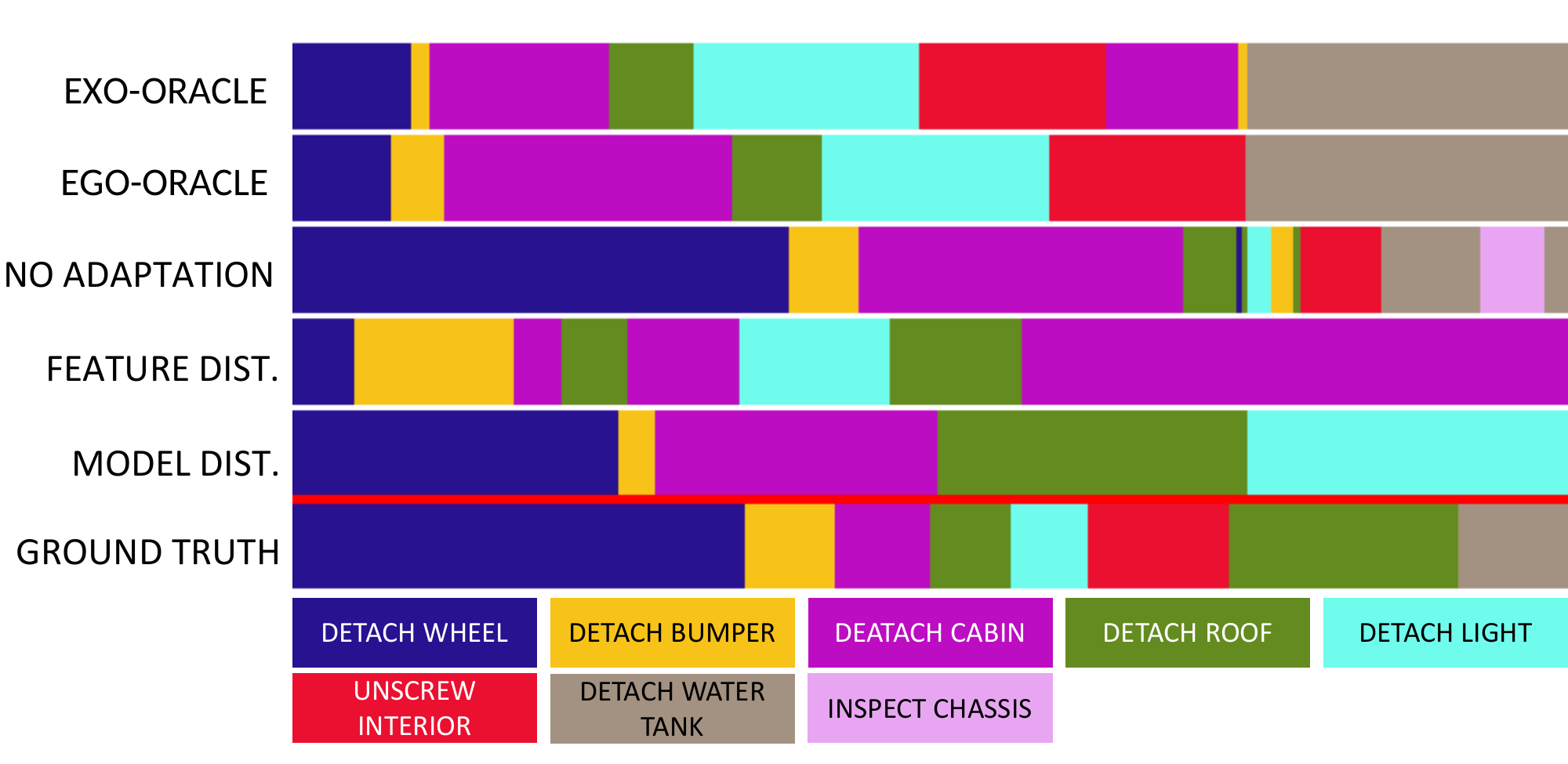}
        \caption{}
        \label{sub:dino_7}
    \end{subfigure}\quad
    \begin{subfigure}{.45\textwidth}
        \includegraphics[width=\textwidth]{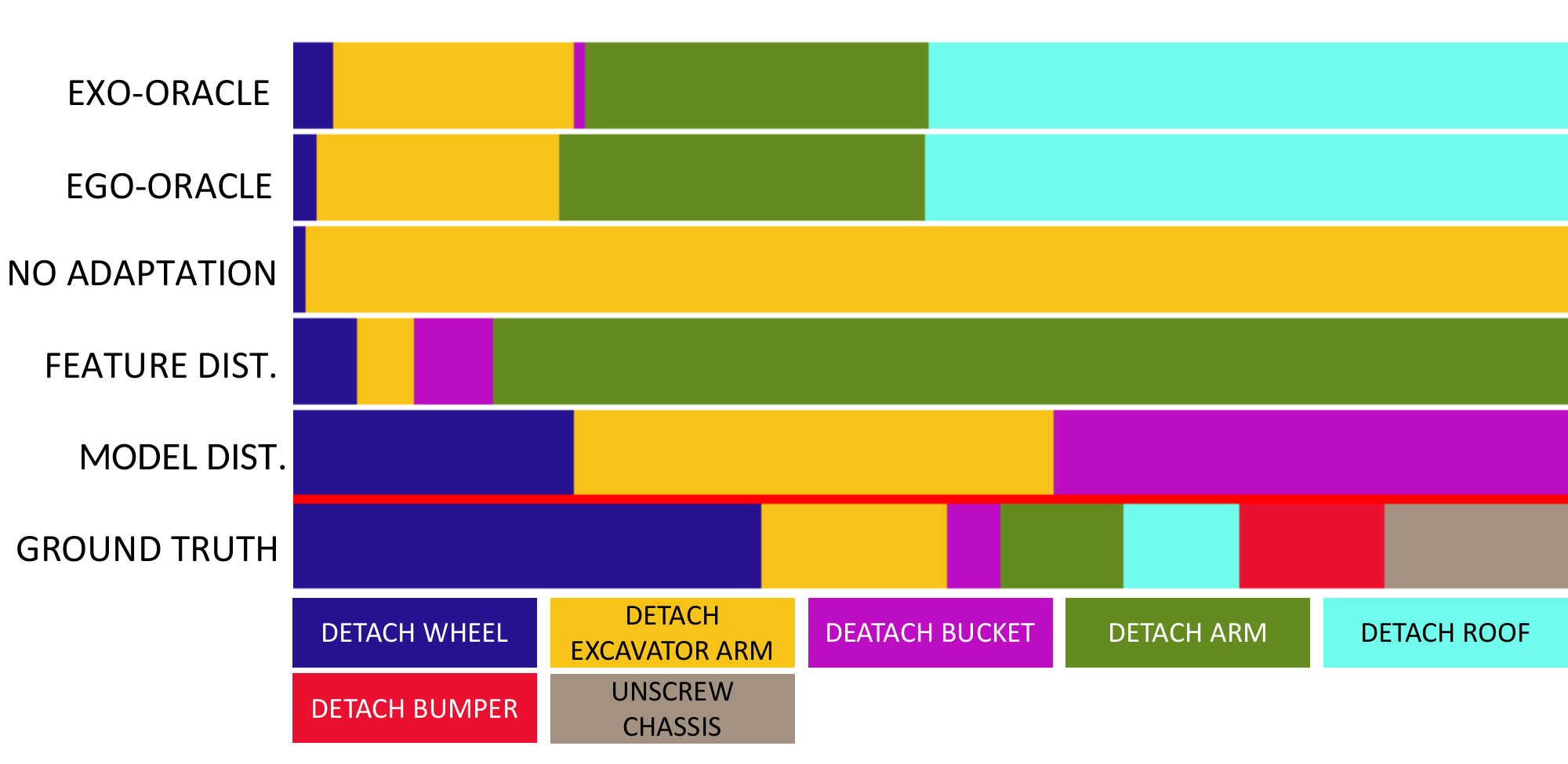}
        \caption{}
        \label{sub:dino_8}
    \end{subfigure}\quad
    \medskip
    \begin{subfigure}{.45\textwidth}
        \includegraphics[width=\textwidth]{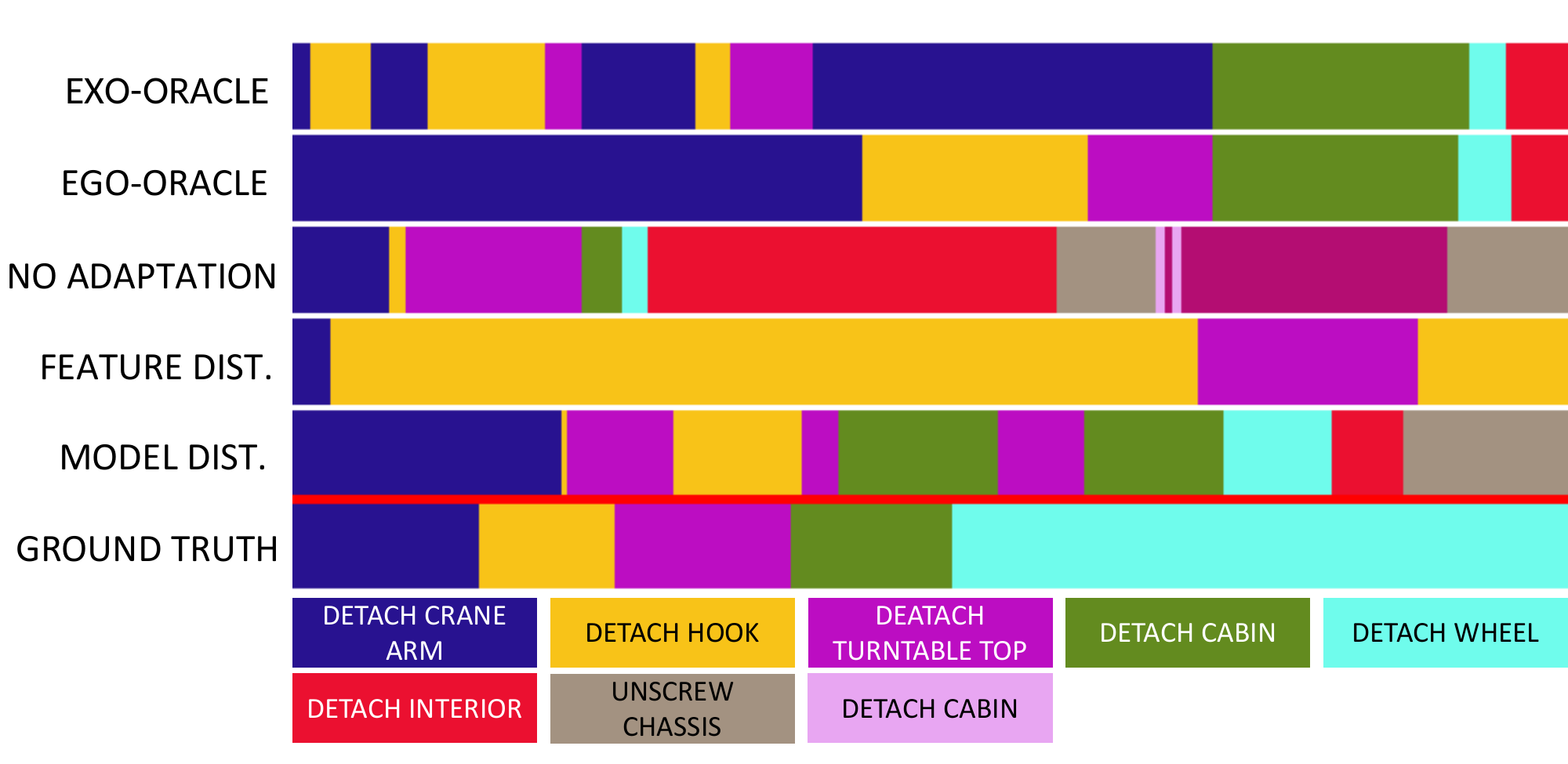}
        \caption{}
        \label{sub:dino_9}
    \end{subfigure}\quad
    \begin{subfigure}{.45\textwidth}
        \includegraphics[width=\textwidth]{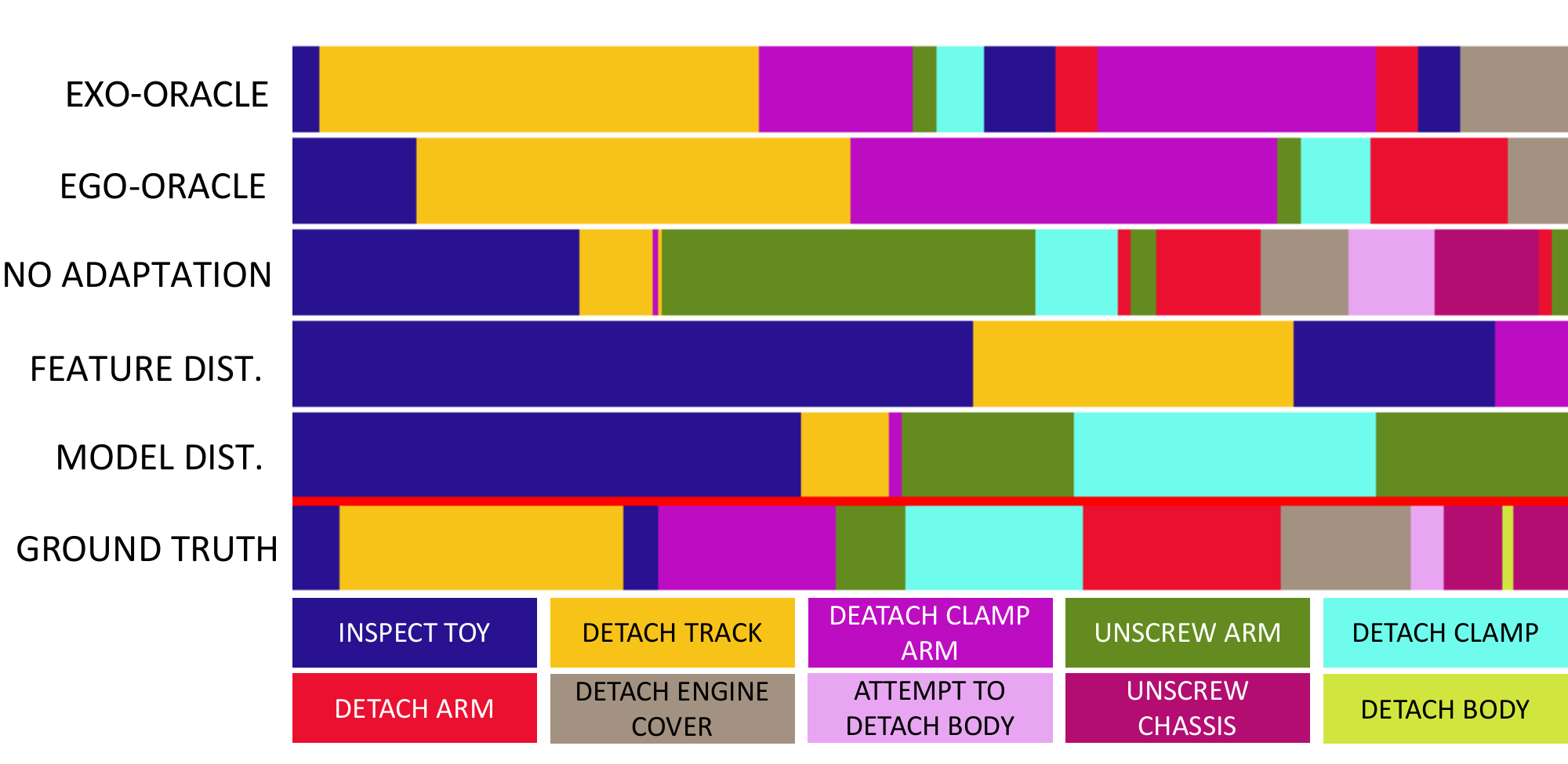}
        \caption{}
        \label{sub:dino_10}
    \end{subfigure}\quad
    \caption{Additional qualitative examples (DINOv2).}
    \label{img:qualitative_supp_DINO}
\end{figure*}

Qualitative results of models based on DINOv2 models (\Cref{img:qualitative_supp_DINO}) highlight how worst performance is obtained by the no-adaptation baseline, which tend to produce over-segmented (i.e.~Figures~\ref{sub:dino_3}, \ref{sub:dino_4}) or under-segmented (Figure~\ref{sub:dino_8}) segmentation.
Exo and Ego oracles generally achieve best performance, usually recovering the correct action sequence (Figure~\ref{sub:dino_2}, \ref{sub:dino_3}, \ref{sub:dino_5}, \ref{sub:dino_7}) even if temporal bounds are not always accurate. Feature distillation often brings some improvement over no-adpatation reducing over-segmentation (Figure~\ref{sub:dino_3}, \ref{sub:dino_4}), but not always recovering the correct order of actions.
TAS model distillation achieves good performance, managing to predict partially correct sequences of actions in different examples (i.e.~Figures~\ref{sub:dino_1}, \ref{sub:dino_2}, \ref{sub:dino_9}).

\subsection{TSM}
\begin{figure*}[t]
    \centering
    \begin{subfigure}{.45\textwidth}
        \includegraphics[width=\textwidth]{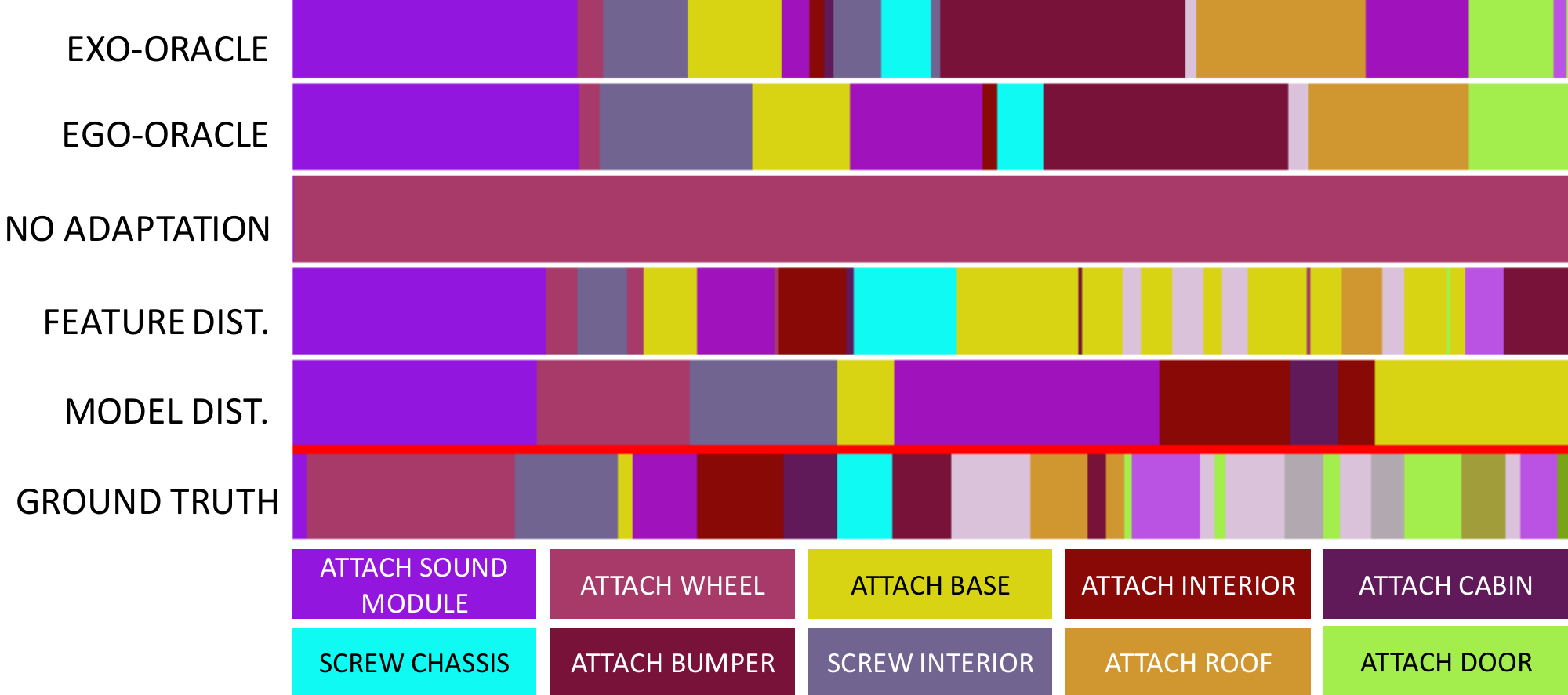}
        \caption{}
        \label{sub:tsm_1}
    \end{subfigure}\quad
    \begin{subfigure}{.45\textwidth}
        \includegraphics[width=\textwidth]{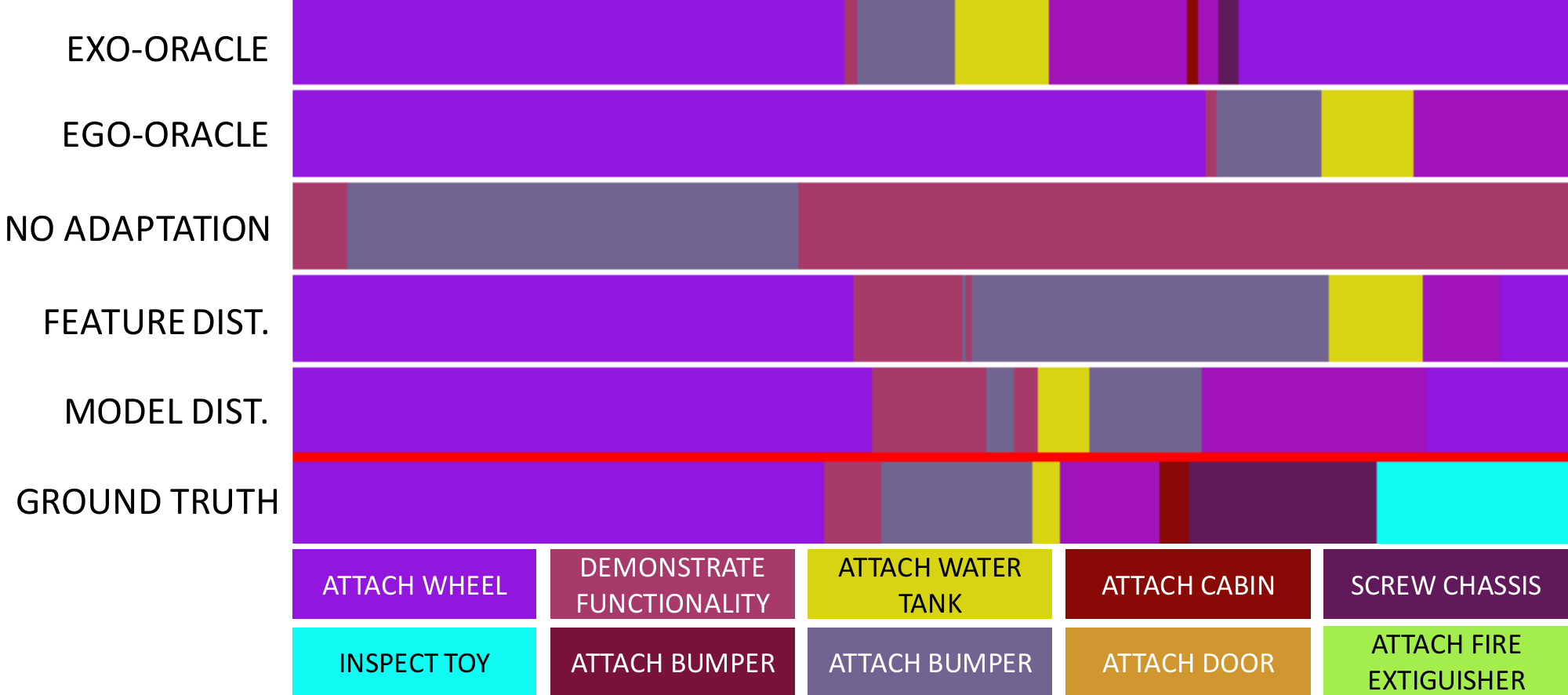}
        \caption{}
        \label{sub:tsm_2}
    \end{subfigure}\quad
    \medskip
    \begin{subfigure}{.45\textwidth}
        \includegraphics[width=\textwidth]{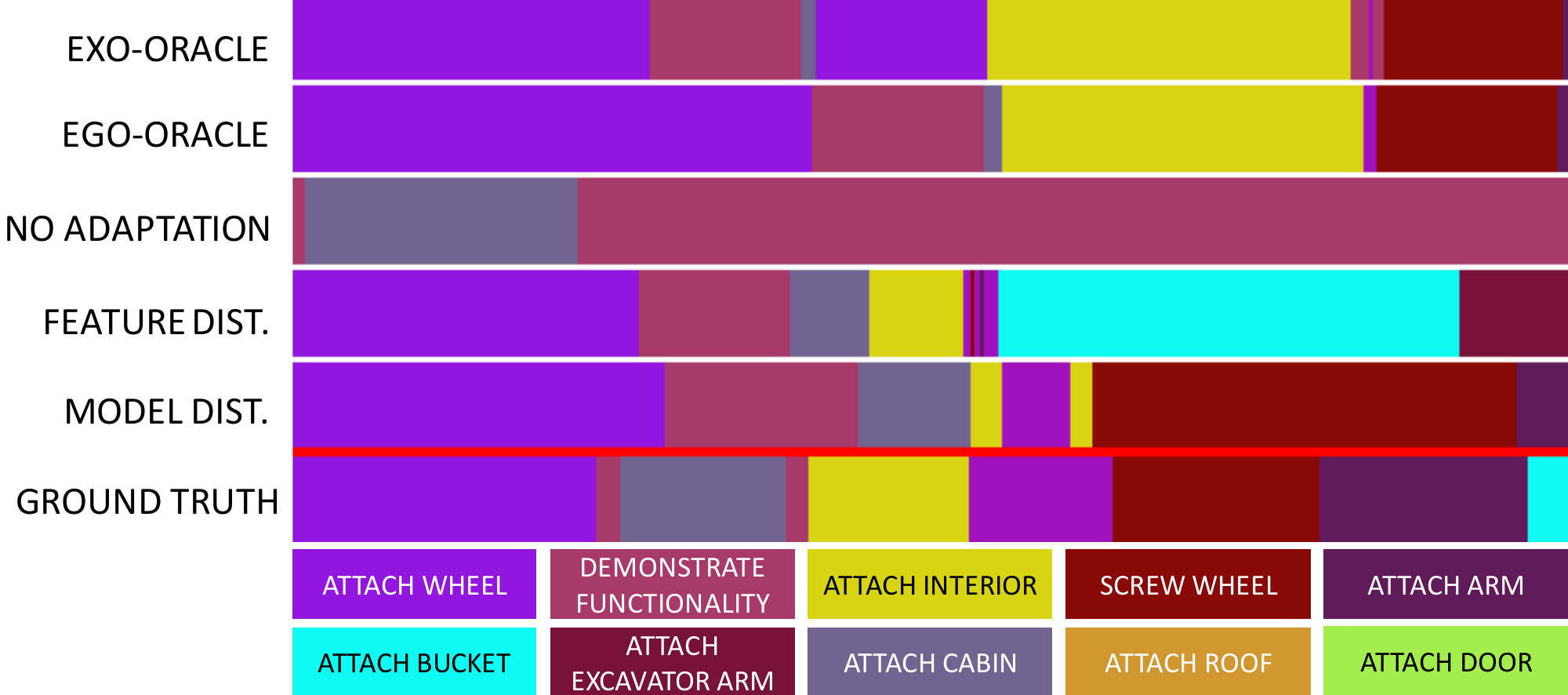}
        \caption{}
        \label{sub:tsm_3}
    \end{subfigure}\quad
    \begin{subfigure}{.45\textwidth}
        \includegraphics[width=\textwidth]{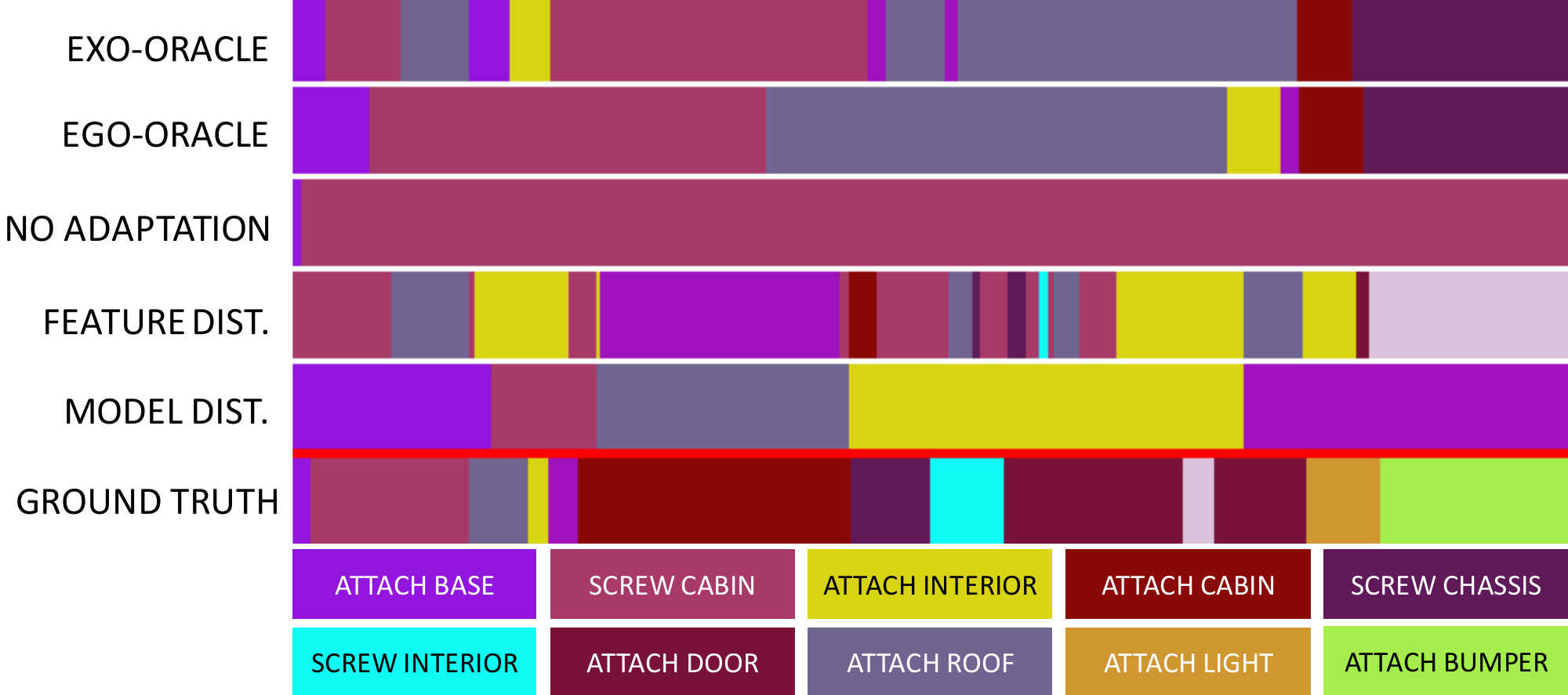}
        \caption{}
        \label{sub:tsm_4}
    \end{subfigure}\quad
    \medskip
    \begin{subfigure}{.45\textwidth}
        \includegraphics[width=\textwidth]{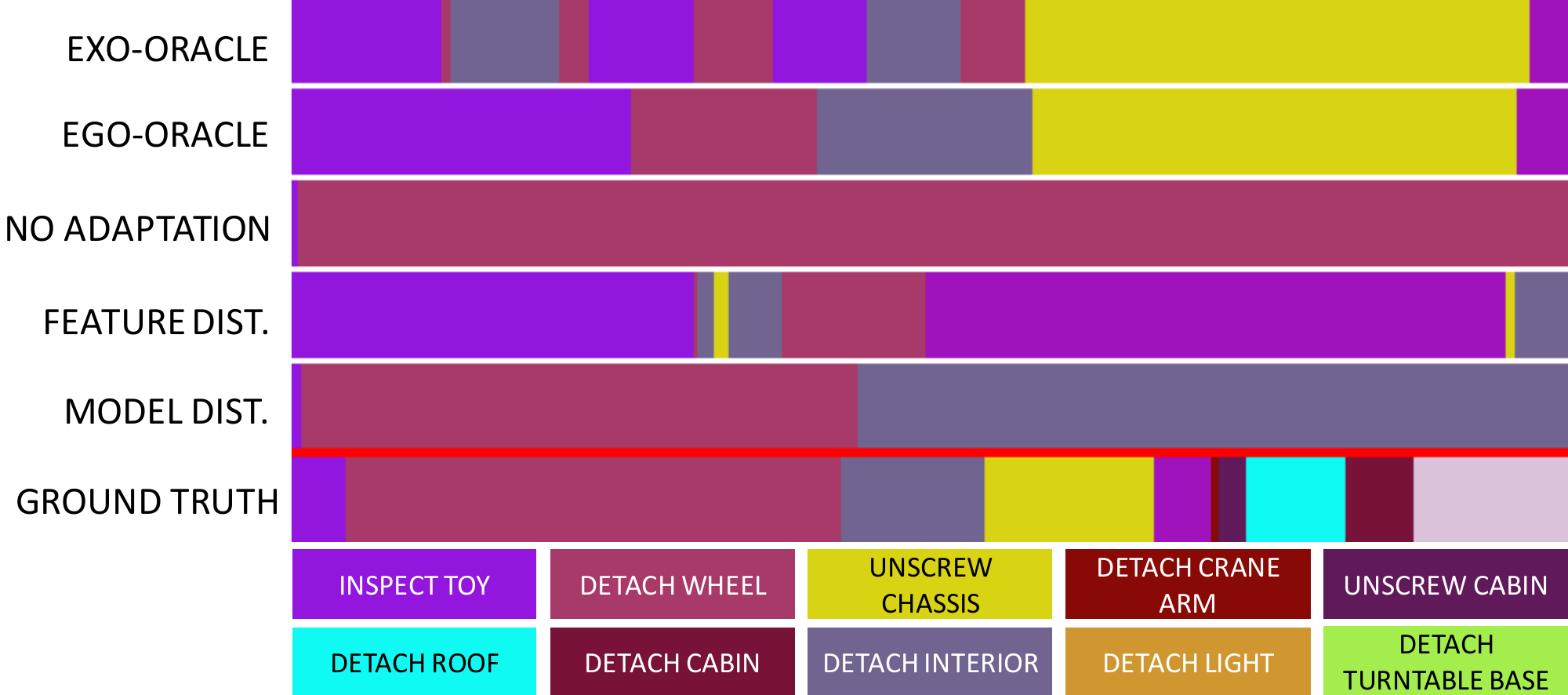}
        \caption{}
        \label{sub:tsm_5}
    \end{subfigure}\quad
    \begin{subfigure}{.45\textwidth}
        \includegraphics[width=\textwidth]{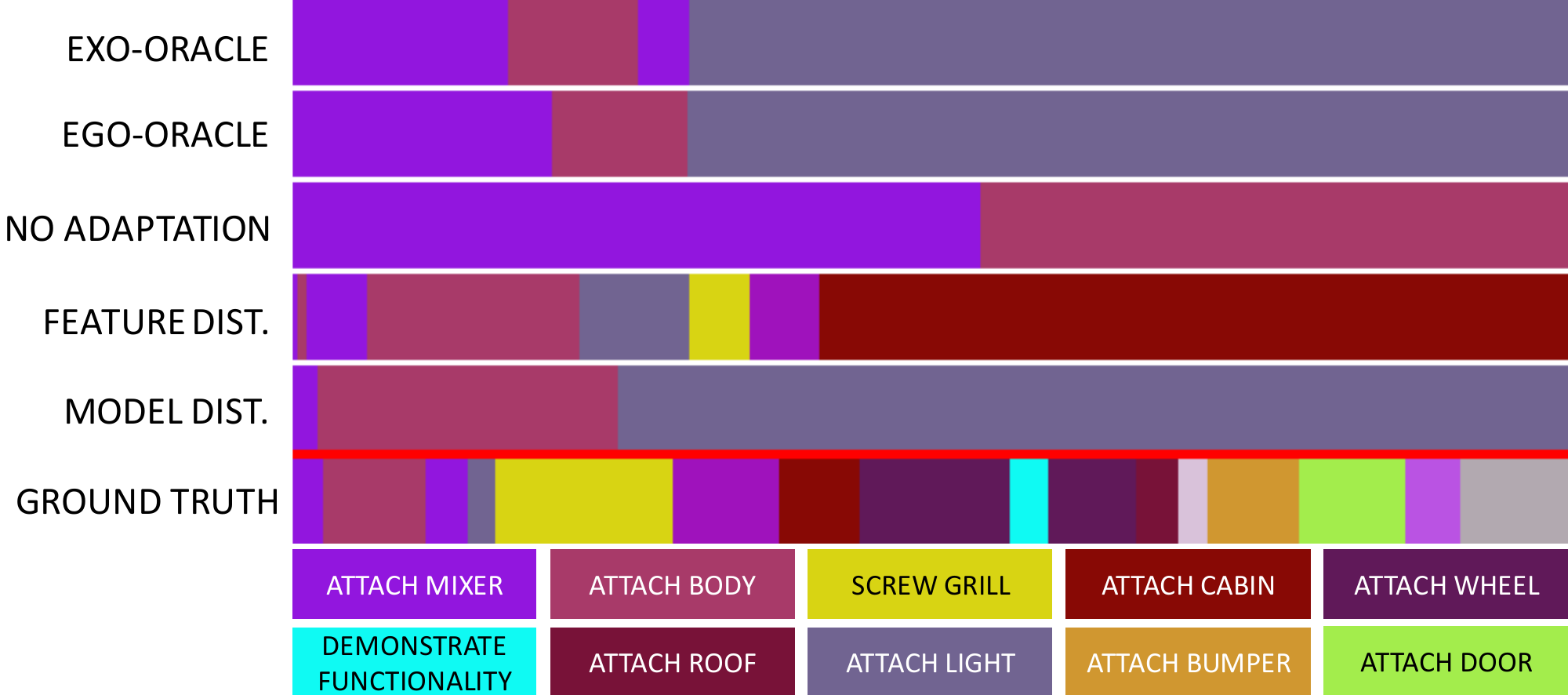}
        \caption{}
        \label{sub:tsm_6}
    \end{subfigure}\quad
    \medskip
    \begin{subfigure}{.45\textwidth}
        \includegraphics[width=\textwidth]{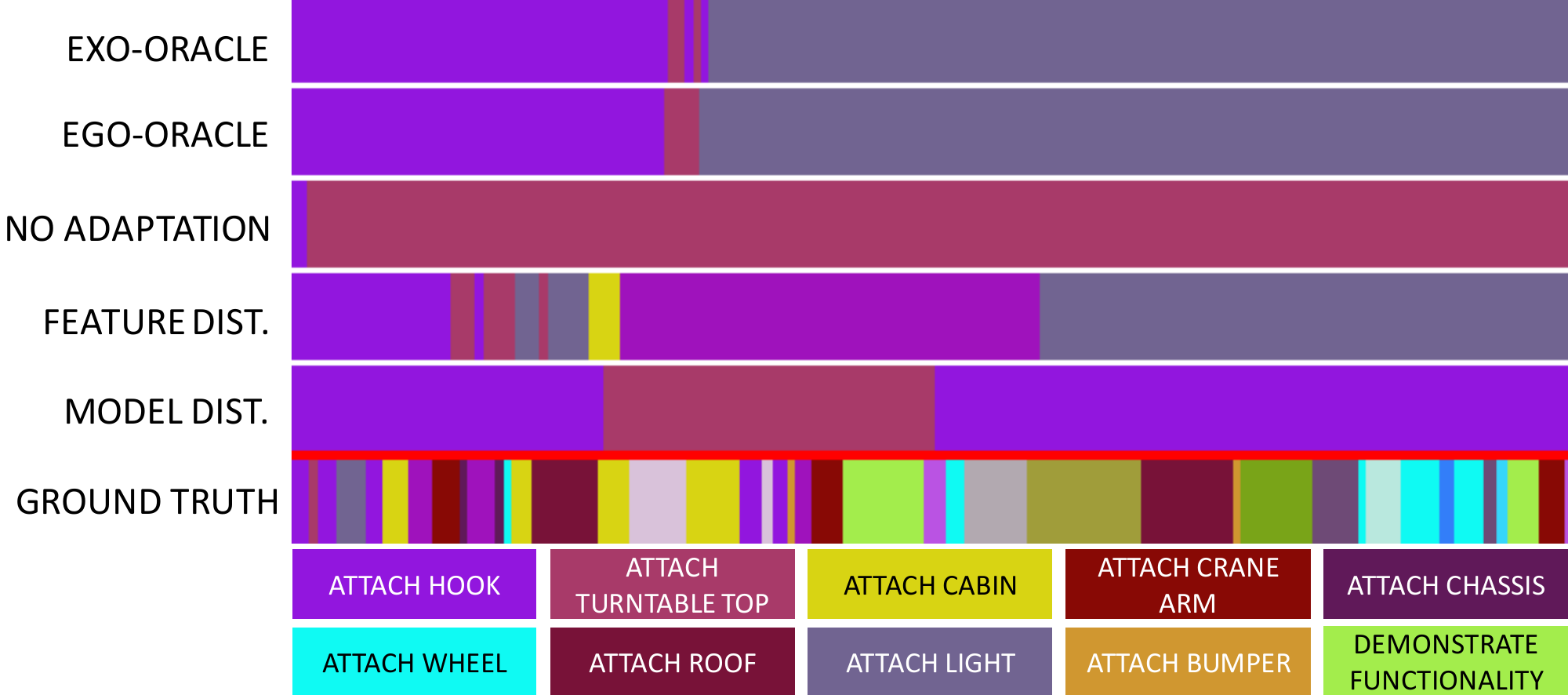}
        \caption{}
        \label{sub:tsm_7}
    \end{subfigure}\quad
    \begin{subfigure}{.45\textwidth}
        \includegraphics[width=\textwidth]{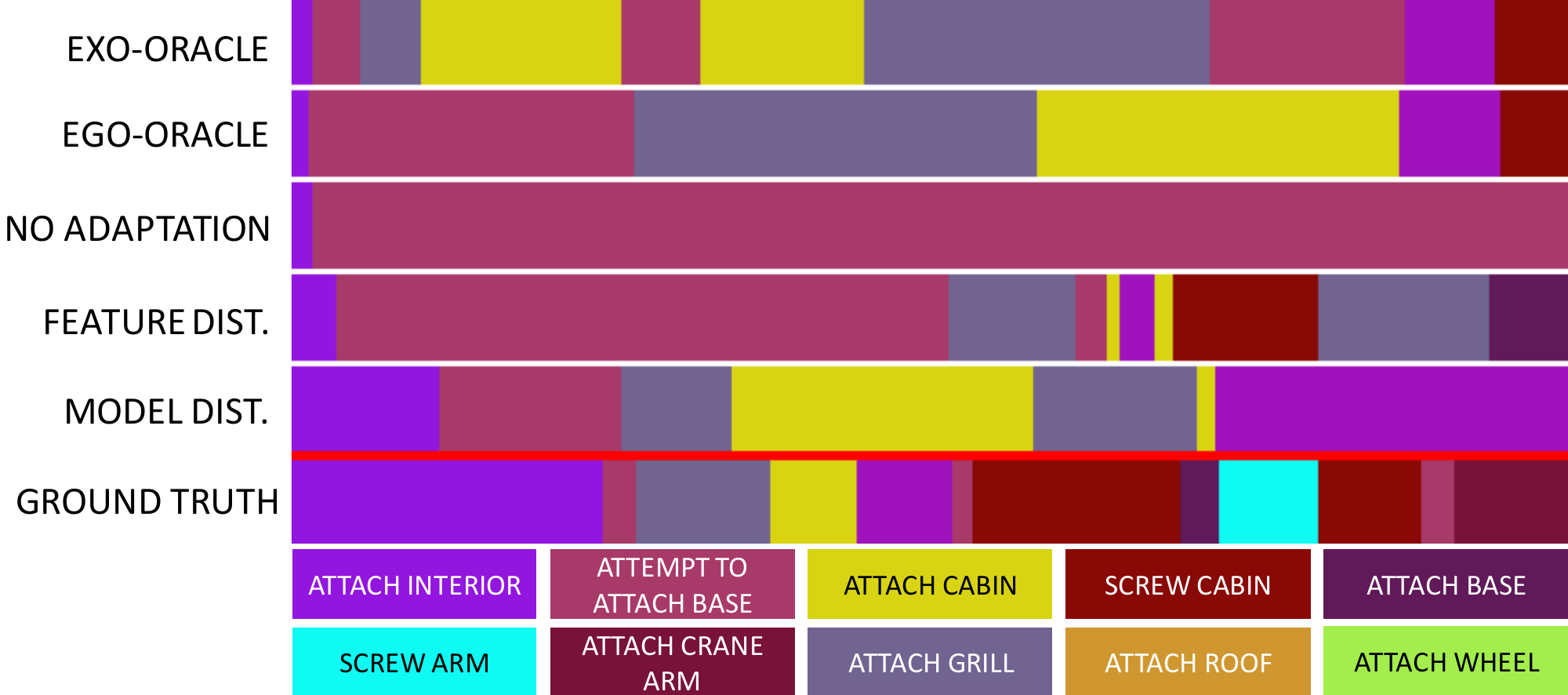}
        \caption{}
        \label{sub:tsm_8}
    \end{subfigure}\quad
    \medskip
    \begin{subfigure}{.45\textwidth}
        \includegraphics[width=\textwidth]{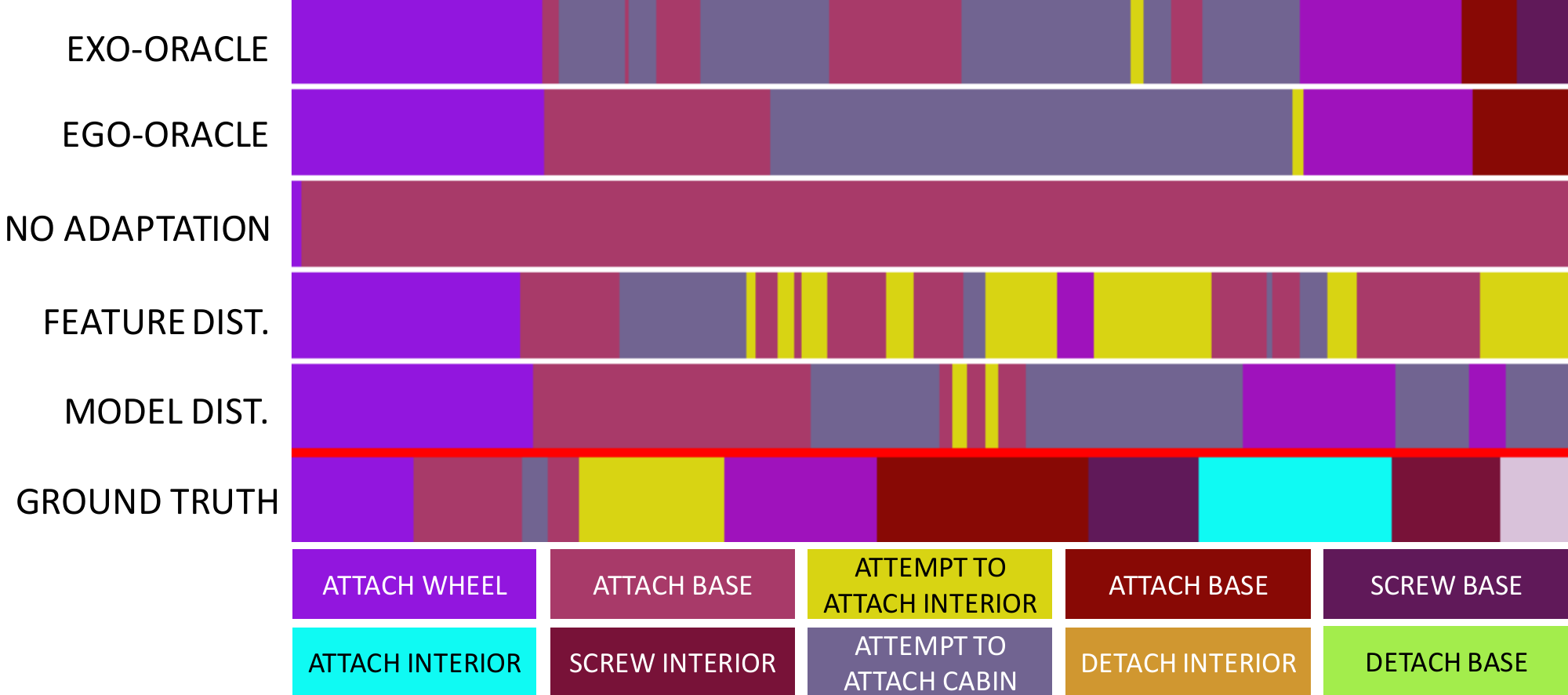}
        \caption{}
        \label{sub:tsm_9}
    \end{subfigure}\quad
    \begin{subfigure}{.45\textwidth}
        \includegraphics[width=\textwidth]{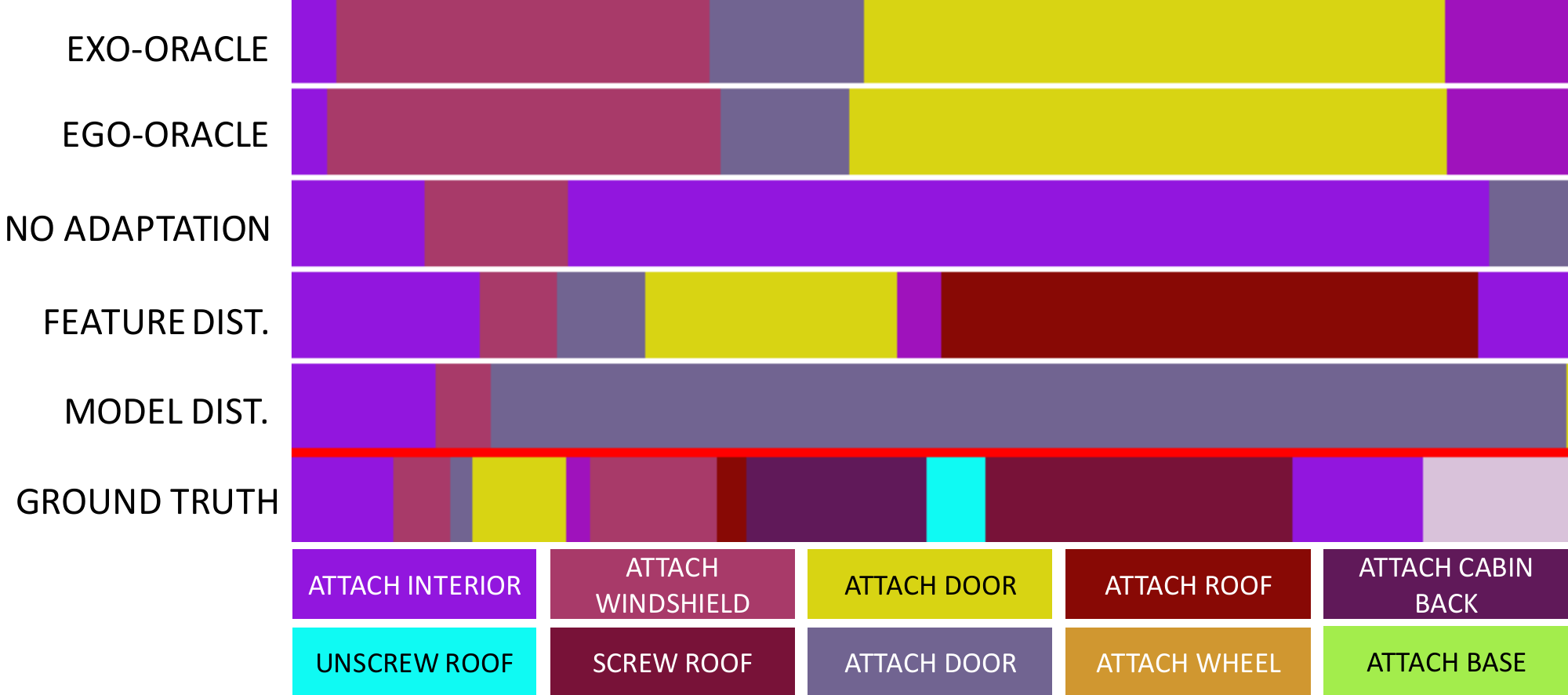}
        \caption{}
        \label{sub:tsm_10}
    \end{subfigure}\quad
    \caption{Additional qualitative examples (TSM).}
    \label{img:qualitative_supp_TSM}
\end{figure*}

Qualitative results of models based on TSM~\Cref{img:qualitative_supp_TSM} bring to similar considerations.
no-adaptation achieve poor results, often leading to under-segmentation (Figure~\ref{sub:tsm_1}, \ref{sub:tsm_5}, \ref{sub:tsm_7}, \ref{sub:tsm_8}).
Oracle baselines often recover partially correct action execution orders (Figure~\ref{sub:tsm_2}, \ref{sub:tsm_5}, \ref{sub:tsm_5}).
Feature distillation improves performance over no-adaptation in many cases, For instance, in the challenging Figure~\ref{sub:tsm_1}-\ref{sub:tsm_6} the under-segmentation of no-adaptation is partially recovered, predicting a correct action ordering for part of the video.
The segmentation produced by feature distillation are in general more accurate (i.e. Figures~\ref{sub:tsm_1}, \ref{sub:tsm_4}, \ref{sub:tsm_10}).

\section{Details on EgoExo4D - View Selection}

\begin{figure*}[!t]
  \centering
   \includegraphics[width=\linewidth]{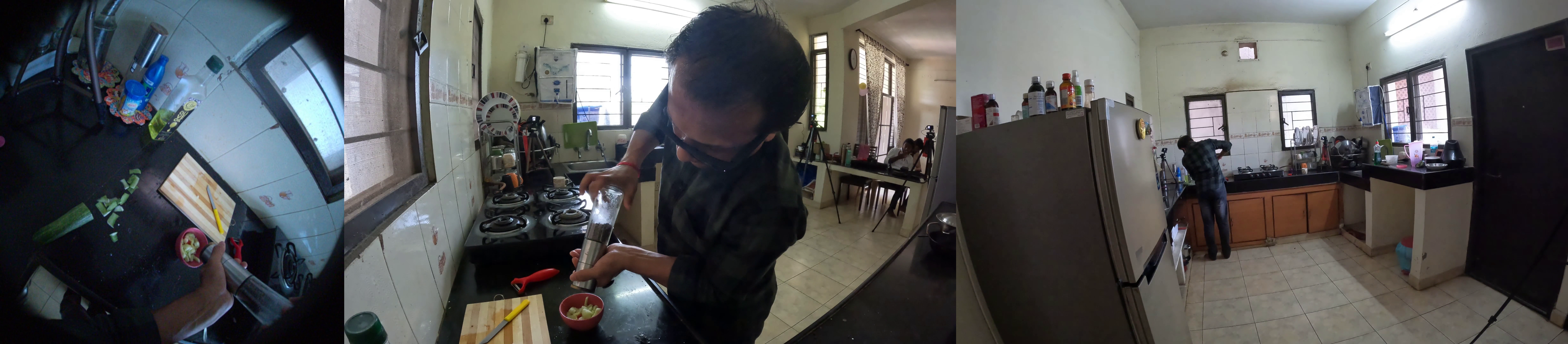} 
   \caption{EgoExo4D views. From left to right: The egocentric view acquired with the Aria glasses, the corresponding exocentric view marked as ``best exo'', another generic camera not marked as best exo. As can be noted, the camera marked as ``best exo'' includes more details on the performed action.}
   \label{fig:egoexo4D_views}
\end{figure*}

As discussed in the main paper, the EgoExo4D dataset was acquired using Aria glasses~\cite{somasundaram2023project} for the egocentric view, with a resolution of $1404\times$1404 pixels, and 4 to 5 GoPro for the exocentric views, with a resolution of $3840\times$2160 pixels. Similar to Assembly101, the egocentric view and the egocentric views are synchronized. Unlike Assembly101, all the cameras used to acquire the dataset are RGB. 
Moreover, since videos are acquired in different environments, fixed cameras are placed differently across videos.
Note that this creates a challenging benchmark in which exo cameras present viewpoint inconsistencies.
We hence selected as the exocentric view for a given video, the one marked by annotators as ``best exo'' for the majority of video segments in that video.
This is the camera which, according to annotators, allows to best understand the performed activity.
When no ``best exo'' annotations are given, we use the camera more frequently marked as ``best exo'' in the dataset.
%For the experiments of the main paper, the exo camera marked by the annotators as best exo was used. However, during the annotation phase, not all recordings contained a camera marked as best exo. To get around this problem, a ranking was drawn up of the cameras that were most often marked as best exo, and the best was used when this information was missing in a recording. 
\Cref{fig:egoexo4D_views} shows an example frame of the dataset.

\section{Details on EgoExo4D - Omnivore Features}

As discussed in the main paper, we adopted Omnivore~\cite{girdhar2022omnivore} features provided by the authors of EgoExo4D.\footnote{\url{https://ego-exo4d-data.org/}} Omnivore is a single model that efficiently classifies images, videos, and single-view 3D data with the same set of model parameters. It takes advantage of transformer-based architectures' flexibility and is trained simultaneously on classification tasks covering various modalities. The common visual representation employed by Omnivore simplifies cross-modal recognition without requiring correspondences between different modalities. The features were extracted using the following setup: features are derived for every take, camera, and camera stream, a stride of $16/30$ seconds is employed, accompanied by a window size of $32/30$ seconds. In cases where the stride does not evenly divide the total duration time, the last [$n - 32/30$, n) seconds of the video are utilized as the final window.

\section{Source of Improvement and role of synchronization}  
%We agree with the reviewer's observation regarding the uncertainty surrounding the source of improvement in our method, i.e. whether the enhancement arises from learning Ego-Exo correspondence via knowledge distillation, which aids in transferring exo knowledge to ego data, or if the improvement can be solely attributed to the larger volume of training data sharing a similar distribution as the test data, or perhaps a combination of both factors. 

%Results suggest that synchronization is crucial for performance: Dropping synchronization (main paper Table~\ref{tab:synchronization}) or reducing the quality of synchronization (Table~\ref{tab:amounts_synchronized}) lead to reduced performance.
To disambiguate the source of improvement (learning correspondences vs volume of training data), in Table~\ref{tab:volumetric}, we compare ``no adaptation'' trained on $10$ hours of labeled exo video wrt our method trained on $5$ hours of labeled exo video and $5$ hours of unlabeled synchronized exo-ego videos. We observe improvements despite methods observe the same number of hours and ours sees less supervised data, suggesting that enhancements are not only due to the increased volume of data. 
To assess if distillation can learn ego-exo correspondences, we run the following experiment: Test exo action segments are represented with TSM-EXO features\footnote{We represent each action segment with the average frame features.} while test ego segments with TSM-EGO features or our distilled features. For each exo action segment, we retrieve the 5 nearest ego segments. When using TSM-EGO features, at least one of the retrieved ego segments has the same action class as the query exo segment $20.52\%$ of the times. This number increases to $32.51\%$ with our distilled features, suggesting that distillation learns ego-exo correspondences.

\begin{table}[t]
\centering

\caption{Results using a fixed amount of hours of training video.}

\begin{tabular}{r|c|c|c|c|c}
 &\textbf{Edit} & $\mathbf{F1_{10}}$ & $\mathbf{F1_{25}}$ & $\mathbf{F1_{50}}$ & \textbf{MoF} \\ \hline
No adaptation & 11.24 & 8.89 & 6.48 & 2.07 & 13.28 \\
Ours  & \textbf{20.31} & \textbf{21.89} & \textbf{17.58} & \textbf{10.54} & \textbf{22.73} \\ \hline
\end{tabular}

\label{tab:volumetric}
\end{table}

\begin{table}[t]
\centering
\caption{Action recognition with TSM features on Assembly101.}

\begin{tabular}{c|c|c|c|c|c}
 &\textbf{Exo-Oracle} & \textbf{Random}&\textbf{Ego-Oracle}& \textbf{No Adaptation} & \textbf{Ours} \\ \hline
Accuracy & 13.4 & 0.5 & 7.15 & 5.10 & 6.65 (+1.55) \\
\hline
\end{tabular}

\label{tab:actionrec}
\end{table}

\section{Generalization to other tasks}
%\AF{Sarebbe utile far vedere risultati di action recognition. Poi direi che comunque è "beyond the scope of the paper" ma abbiamo fatto esperimenti.}
To gain some insights into whether the proposed approach can also be beneficial for other tasks, in Table~\ref{tab:actionrec} we reports results for action recognition on Assembly101, where we observe improvements with repsect to no adaptation. We train MLP models on top of segment-averaged TSM features. This simple model leads to limited absolute performance, but results show that our distillation approach is beneficial also in the case of action recognition.

\end{document}